\documentclass[pmlr]{jmlr}% new name PMLR (Proceedings of Machine Learning)

\RequirePackage{graphicx}
 \usepackage{booktabs}
 \usepackage{multirow}
\usepackage{longtable}% for long tables
\makeatletter
\def\set@curr@file#1{\def\@curr@file{#1}} %temp workaround for 2019 latex release
\makeatother
\usepackage[load-configurations=version-1]{siunitx} % newer version

\theorembodyfont{\upshape}
\theoremheaderfont{\scshape}
\theorempostheader{:}
\theoremsep{\newline}

\jmlryear{2024}
\title[Multi-modal Masked Siamese Network Improves Chest X-Ray Representation Learning]{Multi-modal Masked Siamese Network Improves \\ Chest X-Ray Representation Learning}

\author{\Name{Saeed Shurrab$^*$}
       \Email{saeed.shurrab@nyu.edu}\\ 
       \addr Department of Computer Engineering \\
       New York University Abu Dhabi \\
       Abu Dhabi, United Arab Emirates
       \AND
       \Name{Alejandro Guerra-Manzanares$^*$}
       \Email{alejandro.guerra@nyu.edu}\\ 
       \addr Department of Computer Engineering \\
       New York University Abu Dhabi \\
       Abu Dhabi, United Arab Emirates
       \AND
       \Name{Farah E. Shamout}
       \Email{farah.shamout@nyu.edu}\\ 
       \addr Department of Computer Engineering \\
       New York University Abu Dhabi \\
       Abu Dhabi, United Arab Emirates \\
       \tiny{$^*$These authors contributed equally to this work}
       } 

\begin{document}

\maketitle

\begin{abstract}
Self-supervised learning methods for medical images primarily rely on the imaging modality during pretraining. While such approaches deliver promising results, they  do not leverage associated patient or scan  information collected within Electronic Health Records (EHR). Here, we propose to incorporate EHR data during self-supervised pretraining with a Masked Siamese Network (MSN) to enhance the quality of chest X-ray representations. We investigate three types of EHR data, including demographic, scan metadata, and inpatient stay information. We evaluate our approach on three publicly available chest X-ray datasets, MIMIC-CXR, CheXpert, and NIH-14, using two vision transformer (ViT) backbones, specifically ViT-Tiny and ViT-Small. In assessing the quality of the representations via linear evaluation, our proposed method demonstrates significant improvement compared to vanilla MSN and state-of-the-art self-supervised learning baselines. Our work highlights the potential of EHR-enhanced self-supervised pre-training for medical imaging. The code is publicly available at: \url{https://github.com/nyuad-cai/CXR-EHR-MSN}
\end{abstract}

% and achieves a comparable performance in fine-tuning

% \begin{keywords}
% Self-supervised learning, chest radiographs, masked modeling, electronic health records 
% \end{keywords}

\section{Introduction}
% Introduction to self-supervised learning and learning invariant representations
Supervised training of deep neural networks requires large amounts of quality annotated data~\citep{lecun2015deep}. This is not always straightforward in applications involving clinical tasks, due to the time, cost, effort and expertise required to collect labeled data~\citep{taleb20203d}. Self-supervised learning has recently demonstrated great success in leveraging unlabeled data, such as in natural language processing~\citep{lan2019albert} and computer vision~\citep{jing2020self}. Such frameworks aim to learn useful underlying representations during pretraining, without any labels, which are then used in downstream prediction tasks via supervised linear evaluation.

Considering the state-of-the-art performance of self-supervised pretraining with large unlabeled data compared to end-to-end supervised learning, a plethora of recent applications in healthcare sought to harness the power of self-supervised learning by focusing on a specific type of data, usually a single modality~\citep{shurrab2022self}. For example,~\citet{xie2020pgl} applied spatial augmentations for 3D image segmentation, while~\citet{azizi2021big} applied transformations to Chest X-Ray (CXR) images and dermatology images to predict radiology labels and skin conditions, respectively.~\citet{zhang2022self} preserved time-frequency consistency of time-series data for several tasks, such as detection of epilepsy, while \citet{kiyasseh2021clocs} leveraged electrocardiogram signals to learn patient-specific representations for classification of cardiac arrhythmia. 

In clinical practice, health practitioners rely on several sources of information to perform a diagnosis or interpret medical scans~\citep{cui2021artificial}. For instance, patient sex could be a decisive determinant of diagnostics and therapeutic responses ~\citep{mauvais2020sex}, while patient age could influence clinical decision-making in managing certain health issues~\citep{adams2006influence}. Health practitioners  generally consider additional inputs, such as vital-sign measurements and lab test results, to enhance their understanding of various diseases. Hence, medical data is inherently multi-modal, encompassing various types of modalities, such as medical images, Electronic Health Records (EHR), clinical notes, and omics~\citep{kline2022multimodal}. Consequently, we hypothesize that making use of additional modalities during self-supervised pretraining can improve the quality of the representations for downstream classification tasks~\citep{krishnan2022self}.

Hence, we propose to incorporate EHR data during self-supervised pretraining with a Masked Siamese Network~\citep{assran2022masked} for CXR representation learning. In summary, our pretraining framework includes two visual encoders, adopted from the vanilla MSN, one non-imaging encoder for the EHR, and a projection module that fuses the modalities to encode each CXR-EHR pair. We investigate the inclusion of three types of EHR data, including (a) demographic variables, (b) scan metadata, and (c) inpatient stay information. Our main contributions can be summarized as follows:
\begin{itemize}
\item We design a multi-modal MSN to incorporate EHR data during self-supervised pretraining for CXR, with the goal of improving the quality of learned representations and enhancing downstream classification. This specifically involves the introduction of an EHR encoder and projection head for the multimodal representation of EHR and CXR.
We compare our approach with vanilla MSN, and compare MSN to other state-of-the-art methods for CXR.
\item We conduct a comprehensive evaluation of relevant EHR features, analyzing their individual and combined impact on self-supervised pretraining for downstream classification. %This involves the classification of multiple diseases using both linear evaluation and end-to-end fine-tuning, with a specific emphasis on the low-data regime.
\item We extensively evaluate our proposed approach using three publicly available datasets: MIMIC-CXR \citep{johnson2019mimic} for pretraining and internal validation, and ChexPert \citep{irvin2019chexpert} and NIH-14 \citep{wang2017chestx} for external validation.
\end{itemize}

We summarize the related literature in Section~\ref{related-work}, our proposed methodology in Section~\ref{methods}, the experimental settings in Section~\ref{experiments}, the results in Section~\ref{section:results}, and finally the discussion and concluding remarks in Section~\ref{discussion}.

\section{Related Work}
\label{related-work}
\subsection{Overview of Self-Supervised Learning}
Self-supervised learning methods learn task-agnostic feature representations using hand-crafted pretext tasks or joint embedding architectures \citep{kinakh2021scatsimclr,bardes2021vicreg}. Hand-crafted pretext tasks rely on the use of pseudo-labels generated from unlabeled data. Examples of such tasks include rotation prediction \citep{gidaris2018unsupervised}, jigsaw puzzle solving \citep{noroozi2016unsupervised}, colorization \citep{zhang2016colorful}, and in-painting \citep{pathak2016context}. Joint embedding methods utilize siamese networks \citep{bromley1993signature} to learn useful representations by discriminating between different views of samples based on a specific objective function \citep{bardes2021vicreg}, without the need for human annotation or pseudo-labels. 

Joint embedding methods can be further categorized into contrastive and non-contrastive methods, where the latter encompasses clustering, distillation, and information maximization methods \citep{bardes2021vicreg}. Contrastive methods learn representations by maximizing the agreement between positive pairs and minimizing the agreement between negative pairs \citep{van2018representation}. Some prominent examples include SimCLR \citep{chen2020simple}, contrastive predictive coding \citep{van2018representation}, and MoCo \citep{he2020momentum}. Non-contrastive methods focus on optimizing different forms of similarity metrics across the learned embeddings. Examples include BYOL \citep{grill2020bootstrap}, SimSiam \citep{chen2021exploring}, and VICReg \citep{bardes2021vicreg}. While most existing work considers convolutional networks as backbones for input encoders, recent approaches explore the role of vision transformers (ViT) \citep{dosovitskiy2020image} for self-supervision, such as DINO \citep{caron2021emerging} and MSN \citep{assran2022masked}. MSN is a state-of-the-art self-supervised learning architecture that operates on the principle of mask-denoising, without reconstruction, as well as transformation invariance with transformers. MSN has limited applications in healthcare-related tasks, and is promising considering its computational scalability.

\begin{figure}[t]
    \centering\includegraphics[width=\linewidth,height=7cm]{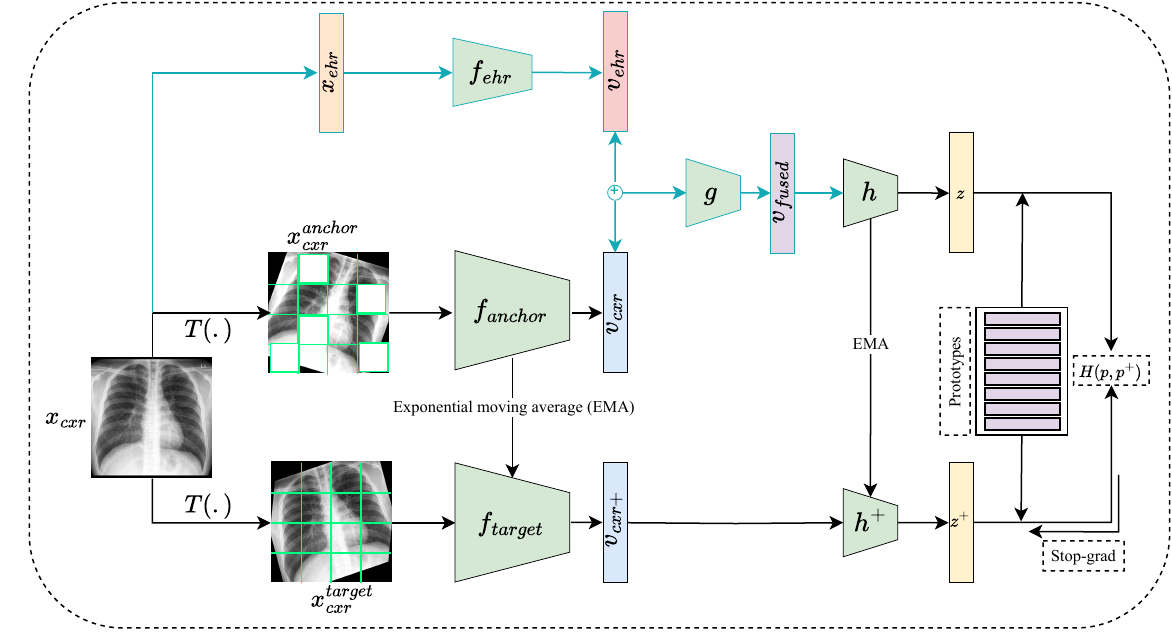}
    \caption{\textbf{Overview of the multi-modal MSN pre-training framework.} We use $x_{ehr}$ as an auxiliary modality during model pretraining. In downstream classification, we freeze $f_{target}$ and use it as a feature extractor along with a classification head for multi-label image classification. Components of the vanilla MSN are outlined in black.} \vspace{-3mm}
    \label{fig:method}

\end{figure}

% \subsection{Self-supervised learning for medical images}
\subsection{Self-Supervised Learning in Healthcare}

Self-supervised learning methods have shown great promise in learning representations of different types of medical images, such as computed tomography and magnetic resonance imaging \citep{jamaludin2017self,zhuang2019self,taleb20203d}, optical coherence tomography and fundus photography \citep{holmberg2020self,hervella2020multi,li2021rotation}, and endoscopy images \citep{ross2018exploiting}. Several studies investigated self-supervised learning for applications involving CXR images. For example,
\citet{pmlr-v143-sowrirajan21a}, \citet{chen2021momentum}, and \citet{sriram2021covid} utilized MoCo \citep{he2020momentum} as a pretraining strategy for chest disease diagnosis and prognosis tasks. \citet{azizi2021big} showed that initializing SimCLR \citep{chen2020simple} during pretraining with ImageNet weights improves downstream performance in CXR classification. \citet{van2024exploring} explored the impact of various image augmentation techniques on siamese representation learning for CXR.

Some studies investigated the use of other sources of information in designing the self-supervised learning framework for CXR, mostly focusing on vision-language pretraining. \citet{vu2021medaug} proposed MedAug, which considers patient metadata for generating positive pairs in MoCo framework \citep{he2020momentum}. \citet{tiu2022expert} achieved expert-level performance by utilizing raw radiology reports associated with CXR as supervisory signals during self-supervised pretraining. Similarly, \citet{zhang2022contrastive} proposed ConVIRT, a self-supervised learning framework that learns representations from paired CXR and textual data. None of these methods explored the incorporation of static EHR data during self-supervised pretraining, nor did they investigate the use of MSN for CXR. To address these gaps, we specifically focus on MSN with the goal of enhancing chest X-ray representation learning using transformer pretraining with EHR data.

\section{Methods}
\label{methods}
\subsection{Problem Formulation}
Consider $x_{cxr}\in \mathbb{R}^{h\times w}$, where $h$ and $w$ represent the image height and width, to be a CXR image collected from patient during their hospital stay, the goal is to predict a set of disease labels $y_{cxr}$. We assume that each $x_{cxr}$ is associated with $x_{ehr} \in \mathbb{R}^{n}$, a vector of static features derived from the patient's EHR data, where $n$ is the number of variables. We use both modalities to learn the CXR representation within our pretraining framework. An overview of the multi-modal MSN architecture for pretraining is shown in Figure \ref{fig:method}. The network consists of two components: (i) the visual encoders for CXR (ii) a multi-modal branch that incorporates the EHR data, as described in the following sections.

\begin{figure}[ht]
    \centering
    \includegraphics[width=0.55\textwidth]{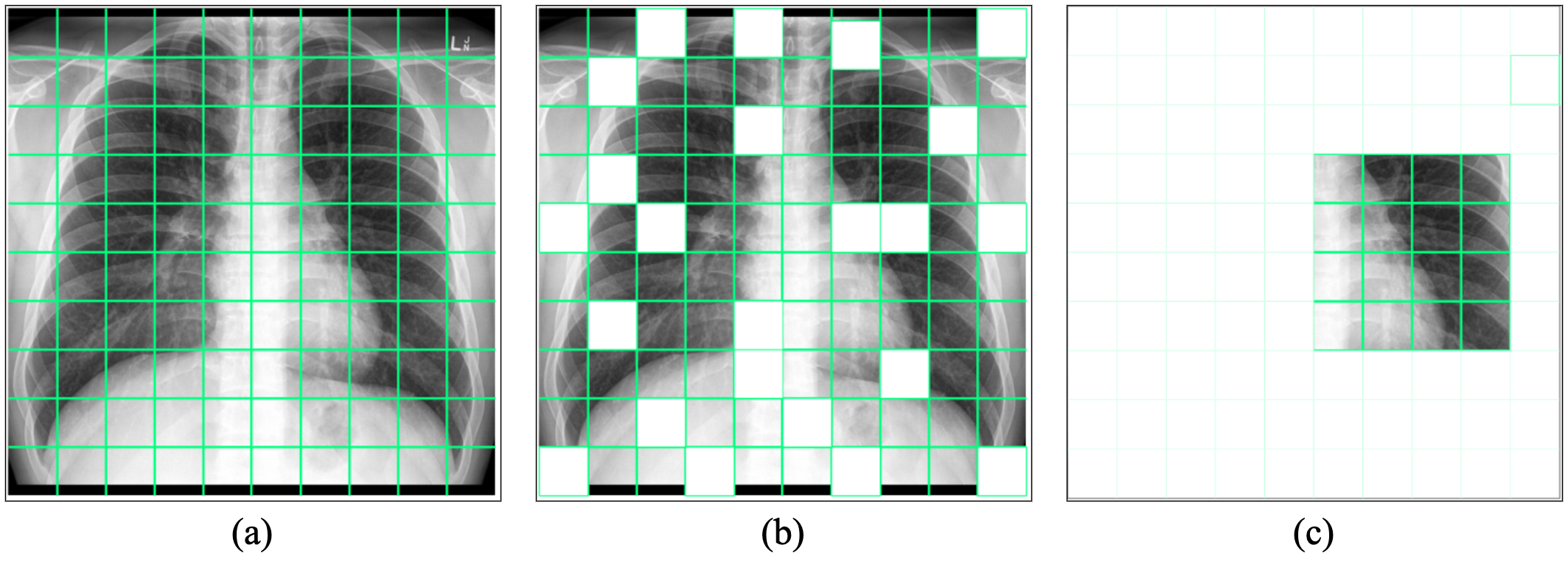}
    \caption{\small\textbf{Visualization of the masking strategies.} (a) Original image. (b) Random masking: drops patches at random locations within the image. (c) Focal masking: keeps a set of neighboring patches and drops the remaining ones.} \vspace{-3mm}
    \label{fig:masking}
\end{figure}

\subsection{Masked Siamese Network}
\label{section:msn}
We adopt the MSN \citep{assran2022masked} as the base model for our proposed framework due to its computational scalability and the need for pretraining transformers efficiently. For a given unlabeled input image, the goal is to align the anchor and target views, denoted by $x^{anchor}_{cxr}$ and $x^{target}_{cxr}$, respectively. For each image, a random transformation function $T(.)$ is used to generate a set of $M$ anchor views and a single target view. The transformations include {image resizing}, {random resized cropping}, and {random horizontal flipping} with a probability of $0.5$, following \cite{pmlr-v143-sowrirajan21a}. However, we excluded color jittering and Gaussian blurring as the former does not apply to grayscale images, while the latter may distort disease-related information \citep{pmlr-v143-sowrirajan21a}. Since the views are patchified to serve as input to a ViT, the anchor views are further masked via patch dropping (either random or focal masking as shown in Figure~\ref{fig:masking}), while leaving $x^{target}_{cxr}$ unmasked.

Two encoders, $f_{anchor}$ and $f_{target}$, parameterized with a ViT \citep{dosovitskiy2020image}, are trained to generate the image embeddings:
\begin{equation}
    v_{cxr} = f_{anchor}(x^{anchor}_{cxr}) \And v_{cxr+} = f_{target}(x^{target}_{cxr}).
\end{equation}

The target embedding $v_{cxr+}$ is further processed by a projection head $h^+$ to obtain $z^+$, while $v_{cxr}$ is used in the multi-modal mapping as described in the next section. MSN does not compute the loss directly on the generated embeddings based on a certain similarity metric. Instead, it utilizes a set of prototypes and seeks to map the projected embeddings of a given sample into the same learned prototype, where the mappings are used to compute the loss. Further background information is provided in Appendix \ref{section:pretrain-setup}.%The MSN loss is further described in Appendix \ref{appendix:msn}.

\subsection{Multi-modal Pretraining}
Instead of solely relying on CXR, our proposed framework encourages MSN to leverage additional information extracted from the patient's EHR. In particular, we introduce three additional modules to the standard MSN architecture. First, we encode the static EHR data with $f_{ehr}$ to learn a representation vector, such that:
\begin{equation}
    v_{ehr} = f_{ehr}(x_{ehr}).
\end{equation}

We then concatenate ($\oplus$) the EHR and CXR latent representations, $v_{ehr}$ and $v_{cxr}$. At this stage, there exists dimensionality mismatch between the concatenated representations and $v_{cxr+}$. To address this, we project the concatenated representation into the same dimensional space of $v_{cxr+}$ using the projection head $g$, resulting in $v_{fused}$:
\begin{equation}
    v_{fused} = g(v_{ehr} \oplus v_{cxr}).
\end{equation}

Lastly, the fused representation is processed by a projection head $h(.)$ to compute $z$. Both $z$ and $z^+$ are used in the clustering of prototypes during pretraining. Overall, this simple procedure learns a joint representation of the EHR data and the anchor embeddings, with the goal of improving the mapping ability of MSN during pretraining.
  
\subsection{Downstream Classification}
After pretraining the encoders $f_{target}$, $f_{anchor}$, and $f_{ehr}$, we use $f_{target}$ as a feature extractor. We then include a classification model $f_c$ to predict the image labels:
\begin{equation}
    \hat{y}_{cxr} = f_c(v_{cxr+})
\end{equation}

The main assumption behind our proposed method is that introducing additional data related to the patient or the CXR scan during pretraining would provide the model with valuable contextual information. This information would contribute to enhancing the quality of the learned representations for downstream classification. 

%\section{Cohort}
\begin{table}[!t]
    \centering
    \caption{\small\textbf{Summary of EHR features incorporated during pretraining.} We pretrain multi-modal MSN using each of the EHR features summarized below.} 
    \resizebox{0.8\linewidth}{!}
    {\begin{tabular}{c c c c c c c} 
\toprule
 \textbf{Group} & \textbf{Feature} &\textbf{Type} & \textbf{Values} & \textbf{Dimension} \\ 
\midrule
 \multirow{2}{*}{\shortstack{Demographic \\ ($x_{D}$)}} & Age & Numeric & $\{ x \in \mathbb{N} | 18 \leq x \leq 100 \}$ & $x_{age} \in \mathbb{R}^{1}$ \\
  & Sex & Binary & $\{Male, Female\}$ & $x_{sex} \in \mathbb{R}^{2}$ \\ \hline
  \multirow{2}{*}{\shortstack{Scan medatadata \\ ($x_{SM}$)}} & View & Multiclass & \shortstack{$\{AP, L, PA, LL\}$} & $x_{view} \in \mathbb{R}^{4}$ \\ 
   & Position & Binary & $\{Erect, Recumbent\}$ & $x_{pos} \in \mathbb{R}^{2}$  \\ \hline
  \multirow{2}{*}{\shortstack{Inpatient stay \\ information ($x_{SI}$)}} & ICU admission & Binary & $\{Negative, Positive\}$ & $x_{icu} \in \mathbb{R}^{2}$   \\ 
   & In-hospital mortality & Binary & $\{Negative, Positive\}$ & $x_{mort} \in \mathbb{R}^{2}$  \\
\bottomrule
\end{tabular}}
\label{tab:pretraining_features}
% \vspace{-4mm}
\end{table}

\begin{table}[t]
    \centering
    \caption{\small\textbf{Summary of datasets and data splits.} We summarize the size of the training, validation, and test sets in terms of the number of images used in our experiments. We used MIMIC-CXR for self-supervised pretraining and downstream classification and CheXPert to obtain an external test set only. For NIH-14, we used the training set during downstream classification since it has different labels from MIMIC-CXR.}
    \resizebox{0.7\linewidth}{!}
    {\begin{tabular}{l c c c c} 
\hline
 \textbf{Dataset} & \textbf{Purpose} &\textbf{Training} & \textbf{Validation} & \textbf{Test} \\ 
\hline
  \textbf{MIMIC-CXR} & Internal validation & 325,188 & 15,282 & 36,625 \\ 
 \textbf{ChexPert} & External validation & - & -  & 688 \\ 
  \textbf{NIH-14} & External validation & 32,457 & 3,567 & 15,735 \\ 
\hline
\end{tabular}}
\label{tab:datasets_splits}
\end{table}

\section{Experiments}
\label{experiments}

\subsection{Dataset}
We conducted our experiments using the MIMIC-CXR dataset, which is a publicly available dataset of CXR images~\citep{johnson2019mimic}. MIMIC-CXR includes 377,110 chest radiographs gathered from 227,835 studies between 2011 and 2016. Each image is associated with a set of ground-truth labels ${y}_{cxr}\in \mathbb{R}^{14}$ that indicates the presence of any of 14 medical conditions extracted from corresponding radiology reports. The medical conditions are \textit{Atelectasis}, \textit{Cardiomegaly}, \textit{Consolidation}, \textit{Edema}, \textit{Enlarged Cardiomediastinum}, \textit{Fracture}, \textit{Lung Lesion}, \textit{Lung Opacity}, \textit{Pleural Effusion}, \textit{Pneumonia}, \textit{Pneumothorax}, \textit{Pleural Other}, \textit{Support Devices}, and \textit{No Finding}. After pre-processing, the dataset consisted of 376,201 individual CXR images related to 65,152 patients and 82,506 stays. 

We used MIMIC-IV \citep{johnson2023mimic}, an EHR database associated with MIMIC-CXR, to extract EHR variables related to the CXR images. Table \ref{tab:pretraining_features} provides a summary of the extracted EHR features, including data types, possible values, and dimensionality. After a thorough investigation of the MIMIC-IV database, we identified three sets of relevant features: patient demographics ($x_D$), scan metadata ($x_{SM}$), and inpatient stay information ($x_{SI}$). For patient demographics, we retrieved patient age ($x_{age})$ for each specific admission and normalized it using \emph{MinMax} normalization, as well as patient sex ($x_{sex}$) encoded as a one-hot vector. For scan metadata, we extracted the scan view ($x_{view}$) and patient position ($x_{pos}$) and encoded them as one-hot vectors. The possible values for view are antero-posterior (AP), lateral (L), postero-anterior (PA), or left-lateral (LL), while for patient position the possible values are erect or recumbent. For inpatient information, we defined $x_{icu}$ and $x_{mort}$ as binary features indicating whether the patient required intensive care treatment or experienced in-hospital mortality, respectively. We provide a summary of the EHR features distributions in Appendix \ref{appendix:datasets}.

We used MIMIC-CXR and MIMIC-EHR for pretraining and downstream classification. We followed the work of~\cite{hayat2022medfuse} to obtain the training, validation, and test splits. The data splits are provided in Table~\ref{tab:datasets_splits}. 

%Figure \ref{fig:distributions} provides the frequency distributions for the EHR features considered in our work during the self-supervised pretraining stage. 

\subsection{External Validation}

 %For consistency, we one-hot encoded all categorical variables.  We categorize the variables into three groups: demographic features ($x_{D}$), consisting of age and sex attributes; scan metadata ($x_{SM}$), which includes view and position features; and inpatient stay ($x_{SI}$) information, encompassing Intensive Care Unit (ICU) admission and in-hospital mortality variables. Note that we use upper case to refer to variable categories, which includes all variables within that category, and lower case to refer to single variables (e.g., $x_{D}$ includes $x_{age}$ and $x_{sex}$). 

We also performed external validation of our pretrained encoders in downstream classification using the ChexPert \citep{irvin2019chexpert} and National Institutes of Health Chest X-Ray Dataset (NIH-14) \citep{wang2017chestx} datasets. ChexPert is a publicly available medical imaging dataset composed of 224,316 chest radiographs of 65,240 patients, collected between 2002-2017, including inpatient and outpatient centers. It includes images and associated radiology reports, which were used to generate labels indicating the presence of 14 diseases. We evaluated the pretrained encoder and classifier obtained with MIMIC-CXR on the ChexPert test set since they have the same labels.

NIH-14 is a publicly available medical imaging dataset composed of 112,120 frontal-view CXR images of 30,805 patients, collected between 1992-2015. Each image is assigned 14 common disease labels. We follow the same split reported in~\cite{irvin2019chexpert} and perform linear evaluation using the NIH-14 training set, since it has different labels from MIMIC-CXR. The data splits are shown in Table~\ref{tab:datasets_splits}.

%A more detailed description of the datasets is provided in Appendix \ref{appendix:datasets}, including a summary of dataset splits (Table~\ref{tab:datasets_splits}).

%\subsection{Electronic Health Records Data}

\subsection{Evaluation Protocol}
\label{section:3.2}

To assess the quality of the learned representations using our proposed pretraining framework, we conducted linear evaluation as the main evaluation protocol following previous work in self-supervised learning~\citep{bardes2021vicreg,garrido2023rankme,huang2022self}. We froze the encoder $f_{target}$ after pretraining, and trained the classification head $f_c$, which is randomly initialized, with embeddings of the whole training set.

 We report performance using the Area Under the Receiver Operating characteristic Curve (AUROC) and the Area Under the Precision-Recall Curve (AUPRC) across all experiments.  We report the 95\% Confidence Intervals (CI) using the bootstrapping method \citep{puth2015variety}. We also performed statistical significance testing by comparing our proposed approach with the best performing baseline. 

%We follow the same protocol for external validation on CheXpert \citep{irvin2019chexpert} and NIH-14 \cite{wang2017chestx} (see Appendices \ref{appendix:datasets}, and \ref{section:ext-val} for a detailed description of the datasets and the external validation setup). 

%\subsection{Downstream Task Evaluation}

%We evaluate our proposed approach on a multi-label classification task using both internal and external downstream task validation. Internal validation is performed on MIMIC-CXR, the same dataset used for pretraining, linear evaluation, and fine-tuning.  

\subsection{Uni-modal Self-supervised Pre-training Baselines}
We compared the results of our proposed method with self-supervised and supervised baselines that are transformer-based for a fair comparison:
\begin{enumerate}
    \item \textbf{Vanilla MSN} \citep{assran2022masked}: MSN is a self-supervised learning model that leverages the idea of mask-denoising while avoiding pixel and token-level reconstruction. We trained the vanilla version following the architecture in the original work. 
    \item \textbf{DINO} \citep{caron2021emerging}: DINO is a transformer-based architecture that learns representation via knowledge distillation from a teacher backbone to a student backbone without labels.
    
    \item \textbf{Masked AutoEncoder (MAE)} \citep{he2022masked}: MAE is a transformer-based encoder-decoder architecture that learns representation by reconstructing the missing pixels within an input image. 
    
    \item \textbf{Supervised}: We train ViT-S and ViT-T in an end-to-end fashion with random and ImageNet \citep{deng2009imagenet} weight initializations.
\end{enumerate}

 In our experiments, we used ViT-Small (ViT-S) as the backbone for our proposed method and the state-of-the-art self-supervised baselines. We also conducted experiments using the ViT-Tiny (ViT-T) backbone. These are smaller versions than the ones defined in \citet{dosovitskiy2020image} and we chose them due to significantly faster training  time with comparable performance compared to the original implementation with ViT-Base (ViT-B). All implementation details and model training settings are provided in Appendix \ref{section:pretrain-setup}.

\begin{table}[t]
    \centering
    \caption{\small\textbf{Performance results for linear evaluation of our proposed method using ViT-S backbone on the MIMIC-CXR dataset}. We summarize the AUROC and AUPRC performance on the test set for the self-supervised methods as well as the 95\% confidence intervals. The best results are shown in blue. We use {\color{green}$\uparrow$} and {\color{red}$\downarrow$} to indicate whether the performance of a given model is 0.5-1.5\% better or worse than the reference model (i.e., vanilla MSN). {\color{green}$\uparrow\uparrow$} and {\color{red}$\downarrow\downarrow$} indicate whether the difference is 1.5\% better or worse than the reference model, respectively. $-$ indicates that the performance difference is less than 0.5\% compared to the reference model.}
    \resizebox{0.6\linewidth}{!}
    {\begin{tabular}{l l l} 
\hline
 %\textbf{Dataset} & \multicolumn{2}{c}{\textbf{MIMIC-CXR}} \\ \midrule
   \textbf{Pretraining} &\textbf{AUROC (CI)} & \textbf{AUPRC (CI)} \\
\hline

ImageNet & 0.703 (0.699, 0.706) \color{red}$\downarrow\downarrow$ & 0.269 (0.266, 0.272) \color{red}$\downarrow\downarrow$ \\
DINO & 0.714 (0.711, 0.718) \color{red}$\downarrow\downarrow$ & 0.278 (0.276, 0.281) \color{red}$\downarrow$ \\
MAE & 0.649 (0.645, 0.653) \color{red}$\downarrow\downarrow$ &  0.223 (0.221, 0.225) \color{red}$\downarrow\downarrow$ \\
MSN & 0.731 (0.727, 0.734)  &  0.291 (0.289, 0.294) \\ \hline

MSN$+ x_{gen}$  & \color{blue}{0.751$^*$ (0.748, 0.754)} \color{green}$\uparrow\uparrow$ & \color{blue}{0.311$^*$ (0.309, 0.314)} \color{green}$\uparrow\uparrow$ \\   
MSN$+ x_{age}$  &  0.746$^*$ (0.743, 0.749)	\color{green}$\uparrow\uparrow$ &  0.307$^*$ (0.305, 0.310) \color{green}$\uparrow\uparrow$ \\
MSN$+ x_{view}$  & 0.747$^*$ (0.744, 0.750) \color{green}$\uparrow\uparrow$	 &  0.307$^*$ (0.305, 0.310) \color{green}$\uparrow\uparrow$ \\
MSN$+ x_{pos}$  &  0.748$^*$ (0.745, 0.752) \color{green}$\uparrow\uparrow$	 &  0.306$^*$ (0.303, 0.309) \color{green}$\uparrow\uparrow$ \\
MSN$+ x_{mort}$  &  0.748$^*$ (0.744, 0.751) \color{green}$\uparrow\uparrow$	 &  0.308$^*$ (0.306, 0.312) \color{green}$\uparrow\uparrow$ \\
MSN$+ x_{icu}$  &  0.746$^*$ (0.742, 0.749) \color{green}$\uparrow\uparrow$	&  0.305$^*$ (0.303, 0.308) \color{green}$\uparrow$ \\ \hline
MSN$+ x_{SD}$ & \color{blue}{0.751$^*$ (0.748, 0.754)} \color{green}$\uparrow\uparrow$ & 0.310$^*$ (0.308, 0.313) \color{green}$\uparrow\uparrow$ \\  
MSN$+ x_{SM}$  & 0.744$^*$ (0.741, 0.747) \color{green}$\uparrow$	 &  0.306$^*$ (0.303, 0.309) \color{green}$\uparrow\uparrow$ \\
MSN$+ x_{IS}$  & 0.749$^*$ (0.746, 0.752) \color{green}$\uparrow\uparrow$ &  0.308$^*$ (0.305, 0.311) \color{green}$\uparrow\uparrow$ \\
MSN$+ x_{SD+SM}$  & 0.742$^*$ (0.739, 0.746) \color{green}$\uparrow$ &  0.302$^*$ (0.300, 0.305) \color{green}$\uparrow$ \\
MSN$+ x_{SD+SI}$  &  0.744$^*$ (0.740, 0.747) \color{green}$\uparrow$	 &  0.306$^*$ (0.304, 0.309) \color{green}$\uparrow\uparrow$ \\
MSN$+ x_{SM+SI}$  & 0.748$^*$ (0.744, 0.751) \color{green}$\uparrow\uparrow$	  &  0.307$^*$ (0.304, 0.310) \color{green}$\uparrow\uparrow$ \\
MSN$+ x_{SD+SM+SI}$  & 0.739$^{\dagger}$ (0.736, 0.743) \color{green}$\uparrow$  &  0.301$^*$ (0.299, 0.305) \color{green}$\uparrow$ \\
\hline
\multicolumn{3}{l}{{\color{black}\small $^*$  Statistical significance results with respect to vanilla MSN ($p < 0.001$)}} \\
\multicolumn{3}{l}{{\color{black}\small $^{\dagger}$  Statistical significance results with respect to vanilla MSN ($p < 0.01$)}} \\
\end{tabular}}
\label{tab:linear-eval}
\vspace{-5mm}
\end{table}

\section{Results}
\label{section:results}
Here, we present the performance results of our proposed method and all self-supervised baselines using the ViT-S. The performance results for ViT-T and all other baselines, including self-supervised and supervised learning methods, are reported in Appendix \ref{section:add-results}.

\subsection{Linear Evaluation Results}

 Table \ref{tab:linear-eval} summarizes the performance results of the linear evaluation experiments using ViT-S. %We defer the results for ViT-T to Table \ref{tab:linear-eval-additional} in Appendix \ref{app:results_linear_evaluation}. 
First, we note that all of the models that incorporate EHR during pretraining achieve a better performance compared to the best performing baseline, vanilla MSN. The improvement is at least $1\%$ in terms of AUROC and AUPRC, with a maximum improvement of $2\%$ when incorporating demographic information, 0.751 AUROC with $x_{sex}$ or $x_{D}$, compared to 0.731 AUROC with MSN. Additionally, in most experiments where we pretrain with a single EHR feature achieve improvements greater than $1.5\%$, while pretraining with groups of features yields slightly lower performance improvements.

Moreover, our proposed strategies demonstrate a significant improvement compared to state-of-the-art self-supervised learning baselines, specifically MAE and DINO, as well as the supervised ImageNet-initialized baseline, reaching an improvement of $10\%$ compared to the worst performing model.
 %However, in this case, despite the performance of the model is worse on the downstream task, the enhancement of the EHR addition is greater, with all variables providing improvements of at least 1.6\% and up to 2.1\% for the AUROC metric. 
 In summary, the results of linear evaluation demonstrate that our approach, which incorporates EHR data during pretraining, enhances the quality of the learned representations and downstream performance in a multi-label disease classification task, surpassing both the vanilla MSN and other baselines. In Appendix~\ref{section:add-results}, we provide further results with ViT-T as the model backbone. In general, we observe similar patterns, demonstrating consistent behaviour with different backbone architectures.

 For the sake of completeness we also report in Appendix~\ref{section:add-results} the results of a secondary evaluation protocol by fine-tuning under low data regimes including $1\%$, $5\%$, and $10\%$ of the full training set. In this setting, we trained both the pretrained encoder $f_{target}$ and the classification head $f_c$ with the training set. We note that the results are on par with vanilla MSN, which is expected since fine-tuning dilutes the effects of incorporating EHR during pretraining.

\begin{table}[t]
    \centering
    \caption{\small{\textbf{Performance results for external validation of our proposed methods using ViT-S backbone under linear evaluation.} We summarize AUROC and AUPRC results on CheXpert and NIH-14 test sets including 95\% confidence intervals. The best results are shown in blue. We use {\color{green}$\uparrow$} and {\color{red}$\downarrow$} to indicate whether the performance of a given model is 0.5-1.5\% better or worse than the reference model (i.e., vanilla MSN).    {\color{green}$\uparrow\uparrow$} and {\color{red}$\downarrow\downarrow$} indicate whether the difference is 1.5\% better or worse than the reference model, respectively. $-$ indicates that the performance difference is less than 0.5\% compared to the reference model.}\vspace{-2mm}}
    \resizebox{\linewidth}{!}
    {
    \begin{tabular}{l | l l | l l}
        \hline
         \multirow{2}{*}{\textbf{Model}} &\multicolumn{2}{c|}{\textbf{CheXpert}} & \multicolumn{2}{c}{\textbf{NIH-14}} \\\cline{2-5}
         &\textbf{AUROC (CI)} & \textbf{AUPRC (CI)} &\textbf{AUROC (CI)} & \textbf{AUPRC (CI)} \\\hline
         MSN & 0.770 (0.755, 0.785) & 0.420 (0.409, 0.453) & 0.699 (0.695, 0.702)  & 0.220 (0.217, 0.226) \\\hline
         
         MSN$+ x_{sex}$   & 0.768 (0.754, 0.788) $-$ & 0.417 (0.414, 0.457) $-$ & 0.734$^*$ (0.730, 0.738) \color{green}$\uparrow\uparrow$&0.265$^*$ (0.261, 0.274)\color{green}$\uparrow\uparrow$ \\
         
         MSN$+ x_{age}$  & 0.772 (0.754, 0.797) $-$ & 0.424 (0.416, 0.446) $-$ & 0.736$^*$ (0.732, 0.739) \color{green}$\uparrow\uparrow$& 0.263$^*$ (0.257, 0.270) \color{green}$\uparrow\uparrow$ \\
         
         MSN$+ x_{view}$ & \color{blue}{0.801$^{\dagger}$ (0.785, 0.825)} \color{green}$\uparrow\uparrow$ & \color{blue}{0.445$^{\dagger}$ (0.420, 0.469)} \color{green}$\uparrow\uparrow$ & 0.732$^*$ (0.728, 0.736) \color{green}$\uparrow\uparrow$& 0.261$^*$ (0.258, 0.270) \color{green}$\uparrow\uparrow$ \\
         
         MSN$+ x_{pos}$  & 0.785 (0.776, 0.798) \color{green}$\uparrow\uparrow$ & 0.426 (0.412, 0.465) \color{green}$\uparrow$ & 0.734$^*$ (0.730, 0.737) \color{green}$\uparrow\uparrow$& 0.261$^*$ (0.257, 0.270) \color{green}$\uparrow\uparrow$ \\
         
         MSN$+ x_{mort}$ & 0.778 (0.743, 0.811) \color{green}$\uparrow$ & 0.424 (0.412, 0.451) $-$ & 0.737$^*$ (0.734, 0.742) \color{green}$\uparrow\uparrow$& 0.265$^*$ (0.261, 0.275) \color{green}$\uparrow\uparrow$ \\
         
         MSN$+ x_{icu}$  & 0.773 (0.739, 0.789) $-$ & 0.422 (0.408, 0.445) $-$ & 0.727$^*$ (0.722, 0.731) \color{green}$\uparrow\uparrow$&0.256$^*$ (0.251, 0.264) \color{green}$\uparrow\uparrow$ \\ \hline
         
         MSN$+ x_{D}$   & 0.776 (0.757, 0.802) \color{green}$\uparrow$ & 0.419 (0.406, 0.439) $-$ & 0.736$^*$ (0.732, 0.740) \color{green}$\uparrow\uparrow$&0.267$^*$ (0.262, 0.276) \color{green}$\uparrow\uparrow$ \\ 
         
         MSN$+ x_{SM}$  & 0.796$^{\dagger}$ (0.775, 0.813) \color{green}$\uparrow\uparrow$ & 0.427 (0.418, 0.473) \color{green}$\uparrow$ & 0.732$^*$ (0.728, 0.735) \color{green}$\uparrow\uparrow$& 0.258$^*$ (0.254, 0.268) \color{green}$\uparrow\uparrow$ \\
         
         MSN$+ x_{SI}$   & 0.771 (0.753, 0.788) $-$ & 0.423 (0.402, 0.443) $-$ & \color{blue}{0.738$^*$ (0.734, 0.741)} \color{green}$\uparrow\uparrow$& \color{blue}{0.270$^*$ (0.264, 0.278)} \color{green}$\uparrow\uparrow$ \\
         
         MSN$+ x_{D+SM}$   & 0.776 (0.756, 0.791) \color{green}$\uparrow$ & 0.412 (0.397, 0.447) \color{red}$\downarrow$ & 0.728$^*$ (0.725, 0.732) \color{green}$\uparrow\uparrow$& 0.254$^*$ (0.250, 0.263) \color{green}$\uparrow\uparrow$ \\
         
         MSN$+ x_{D+SI}$   & 0.757 (0.724, 0.769) \color{red}$\downarrow$ & 0.418 (0.404, 0.441) $-$ & 0.728$^*$ (0.724, 0.732) \color{green}$\uparrow\uparrow$& 0.269$^*$ (0.263, 0.279) \color{green}$\uparrow\uparrow$ \\
         
         MSN$+ x_{SM+D}$   & 0.782 (0.761, 0.803) \color{green}$\uparrow$ &0.431 (0.418, 0.452) \color{green}$\uparrow$ &0.734$^*$ (0.730, 0.738) \color{green}$\uparrow\uparrow$&0.263$^*$ (0.258, 0.273) \color{green}$\uparrow\uparrow$ \\
         
         MSN$+ x_{D+SM+SI}$   & 0.775 (0.753, 0.798) \color{green}$\uparrow$ & 0.427 (0.413, 0.452) \color{green}$\uparrow$ & 0.725$^*$ (0.721, 0.728) \color{green}$\uparrow\uparrow$& 0.251$^*$ (0.248, 0.258) \color{green}$\uparrow\uparrow$ \\
         \hline
        \multicolumn{5}{l}{{\color{black}\small $^*$  Statistical significance results with respect to vanilla MSN ($p < 0.001$)}} \\
        \multicolumn{5}{l}{{\color{black}\small $^{\dagger}$  Statistical significance results with respect to vanilla MSN ($p < 0.01$)}} \\
    \end{tabular}}
    \label{tab:ext-val-t-s}

\end{table}

\subsection{External Validation Results}
Table \ref{tab:ext-val-t-s} presents the external validation results for linear evaluation of our proposed methods using the ViT-S backbone on the ChexPert \cite{irvin2019chexpert} and NIH-14 \cite{wang2017chestx} datasets.  
%We perform external validation of our proposed models for both self-supervised techniques on the same downstream task and both model architectures. We use the test sets of ChexPert and NIH-14 for that purpose.
The results indicate that our multi-modal pretraining enhances the robustness of self-supervised models, resulting in higher validation scores and improved performance compared to vanilla MSN in most scenarios. In the best scenario, the performance gain reaches $3$-$4\%$ in both evaluation datasets. However, we observe better performance improvements in the NIH-14 dataset, since its training set was used to train the linear evaluation classifier, compared to off-the-shelf evaluation with CheXpert. The results for the VIT-T backbone and fine-tuning are reported in Appendix~\ref{section:add-results}. We observe similar patterns that are in line with the findings of internal validation with MIMIC-CXR.  %In this case, we observe that, in general, our approach performs on par to on both datasets, thus not affecting negatively to its generalization. 

\begin{figure}[ht]
    \centering
    \includegraphics[scale=0.4,width=0.95\linewidth]{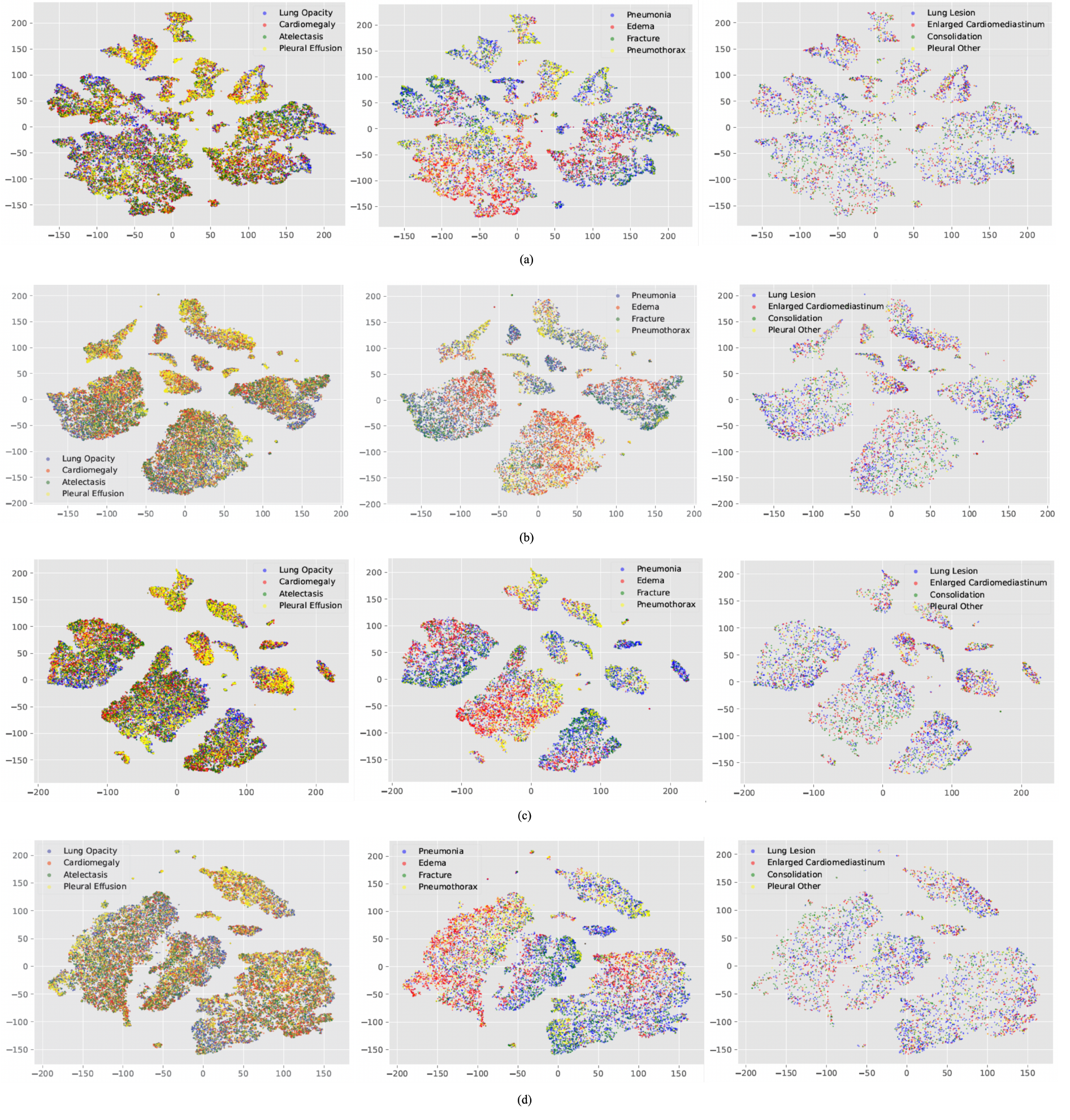} \vspace{-3mm}
    \caption{\textbf{Embedding visualization via t-SNE.} We plot the resulting embeddings obtained via t-SNE for (a): Vanilla MSN (best performing baseline) (b): MSN$+ x_{sex}$ (best performing proposed variant) (c):  MSN$+ x_{D}$  (best performing proposed variant) (d): MSN$+ x_{D+SM+SI}$  (worst performing proposed variant). The first column depicts the clusters of the four most prevalent diseases in the dataset, the second column corresponds to the second four most prevalent diseases, and the last column represents the four least prevalent diseases. 
    } \vspace{-3mm}
    \label{fig:t-sne}
\end{figure}

\subsection{t-SNE Analysis Results}

% \label{appendix_t-sne_results}

 We applied t-SNE to a subset of samples with a single label in the training set (n=68,000) of MIMIC-CXR for the ease of interpretability, considering the challenge of visualizing multi-label samples in the latent space. Figure \ref{fig:t-sne} visualizes the embeddings obtained with vanilla MSN (baseline), MSN$+ x_{sex}$, MSN$+ x_{D}$, and MSN$+ x_{D+SM+SI}$ for the ViT-S backbone. Based on our qualitative assessment, we observe that the clustering quality is enhanced using MSN$+ x_{sex}$, MSN$+ x_{D}$ and MSN$+ x_{D+SM+SI}$ compared to vanilla MSN. More specifically, the clusters of the best performing approaches (MSN$+ x_{sex}$, and MSN$+ x_{D}$) show denser clusters, with lower intra-cluster distance and larger inter-cluster distances, which is typically related to a better clustering quality in the embedding space. We hypothesize that this enhanced clustering quality yields an improved downstream performance.

% Table \ref{tab_appendix:ext-val-t-s} reports the results for fine tuning. and Table \ref{tab:ext-val-t-2}

%\newpage
%\vspace{900mm}
\section{Discussion}
\label{discussion}
In this paper, we propose the incorporation of EHR data during self-supervised pretraining for CXR representation learning, with the aim of improving performance in downstream classification. We present our proposed approach as an enhanced version of an existing self-supervised learning method, MSN \citep{assran2022masked}, by including clinical imaging-related patient EHR data during pretraining. We also comprehensively evaluated our proposed approach on the MIMIC-CXR dataset (internal validation), and two datasets for external validation, ChexPert and NIH-14. Our results demonstrate that the integration of EHR metadata during pretraining with MSN significantly improves performance in the downstream linear evaluation protocol. We also observe that, in general, the inclusion of a single EHR feature yields better quality embeddings compared to pretraining with combinations of EHR features, as shown by the t-SNE embeddings in Figure~\ref{fig:t-sne}. In addition, our proposed method shows superior performance compared with other state-of-the-art self-supervised baselines, including DINO \citep{caron2021emerging} and MAE, \citep{he2022masked} as well as fully supervised baselines pretrained with ImageNet \citep{deng2009imagenet}. Our extensive evaluation highlights that our proposed approach generalizes well to external validation datasets, significantly outperforming vanilla MSN in various settings for CheXpert and in all cases for the NIH-14 dataset. 

A relevant insight derived from our extensive experimentation using single features and combinations of them is that it is \emph{not guaranteed that the addition of more EHR features leads to improved performance}. We hypothesize that such behavior could be attributed to feature interactions, which is an area of future work. %Another possible cause of such behavior is the sparsity of the EHR vector under feature combinations, which may hurt the quality of the learned representations. A thorough investigation of the causes behind this behavior is out of the scope of the present paper and will be addressed in our future work. 

  %We hypothesize that the more balanced distributions shown by these features may help the model to learn better representations, as evidenced by the enhanced clustering of the t-SNE embeddings provided in . In this regard, we observe that all features (MSN+$x_{D+SM+SI}$) did not enhance clustering to the same degree as some individual variables (e.g., MSN+$x_{sex}$) or groups (e.g., MSN+$x_{D}$) did. As a result, downstream performance was superior for those models with better clustering quality, such as MSN+$x_{sex}$ and MSN+$x_{D}$. 

\paragraph{Limitations}
Despite the performance improvements shown by our framework during linear evaluation, our approach has some limitations. First, our approach did not show significant improvements during end-to-end fine-tuning. That is, it performed on-par with vanilla MSN, or slightly lower in the worst scenario as shown in the additional results in  Appendix~\ref{section:add-results}. We hypothesize that fine-tuning the target encoder without the EHR data dilutes the impact of incorporating it during pretraining. Specifically, the learned representations during the pretraining stage (with EHR data) are updated during fine-tuning (without EHR data), which impacts the quality of the representations.  We aim to address this limitation in our future work, where the EHR data is incorporated during both linear evaluation and fine tuning, rather than solely during pre-training for a fair comparison.  Furthermore, we only tested our proposed methodology with a transformer-based self-supervised learning method. In future work, we will explore its applicability to other self-supervised learning architectures and generalization as a framework for other medical imaging datasets and clinical prediction tasks.

%Another limitation is that our proposed approach has only been tested using medical imaging datasets associated with EHR databases, such as MIMIC-CXR and MIMIC-IV. This restricts its applicability to datasets consisting of medical images.

% ACKNOWLEDGEMENTS ONLY GO IN THE CAMERA-READY, NOT THE SUBMISSION
% \acks{Many thanks to all collaborators and funders!}

%Do NOT change font size of references or modify the bibliography style
\bibliography{main}

\newpage
\appendix

\setcounter{table}{0}
\renewcommand\theHtable{Appendix.\thetable}
\renewcommand{\thetable}{A\arabic{table}}

\section{Implementation details}

In this section, we present additional information pertaining to the implementation details.

\label{section:pretrain-setup}

\subsection{Masked Siamese Network}
The MSN network \citep{assran2022masked} is a self-supervised pretraining architecture that leverages mask de-noising and transformation invariance. Let $B\geq1$ be a mini-batch of chest X-ray images sampled from an unlabeled training set. For each image in the dataset, a set of random transformations is applied to obtain $M$ anchor views $x_{i,1}, x_{i,2}...x_{i,M}$ and a single target view denoted as $x_i^+$. As the images serve as input to a Vision Transformer (ViT), the anchor and target views are initially \emph{patchified}, i.e., converted into a set of patches with dimensions $p\times p$ pixels, where $p$ represents the patch size. Before the patches are processed by the transformer-based encoders, random and focal masking are applied to the anchor views. Figure \ref{fig:masking} illustrates the masking strategies: random masking drops random patches at different locations across the input image, while focal masking drops a block of adjacent patches. %In our setup, we use a random masking ratio of 0.3 (or 30\%) for MSN pretraining, which is equivalent to dropping $(224 \div 16)^2 \times 0.3 \approx 59$ patches out of the 196 patches. 

The anchor patches are processed by the encoder $f_{anchor}$, while the target patches are processed by the encoder $f_{target}$ (see Figure \ref{fig:method}). The weights of $f_{target}$ are updated through an exponential moving average of $f_{anchor}$. Both encoders are parameterized as ViT encoders. The [CLS] token of each encoder is processed by a three-layer projection head $h$ to obtain the final image embeddings $z_{i,m}$ and $z_{i}^{+} \in \mathbb{R}^{n}$ for each encoder.

There also exists a set of learnable prototypes $\mathbf{q}\in \mathbb{R}^{n\times k}$, where $K$ is the number of prototypes. The network learns to assign the learned embeddings to the prototypes based on cosine similarity, such that $p_{i,m} := \text{softmax}(\frac{z_{i,m} \cdot \mathbf{q}}{\tau})$ and $p_{i}^+ := \text{softmax}(\frac{z_{i}^+ \cdot \mathbf{q}}{\tau^+})$, where $\tau$ and $\tau^+$ are the temperature hyperparameters. As described in \citet{assran2022masked}, the overall loss function is defined using a standard cross-entropy loss $H(p_i^+, p_{i,m})$ and a mean entropy maximization regularizer $H (\bar{p})$ following Equation \ref{msn:loss}:

\begin{equation}
      \label{msn:loss}
      \mathcal{L} =  \frac{1}{MB}  \sum^{B}_{i=1}\sum^{M}_{m=1}H(p^{+}_{i}, p_{i,m}) - \lambda H (\bar{p}),
\end{equation}

where $B$ is the batch size, $M$ is the number of anchor views, and $\lambda$ is the weight of the mean entropy maximization regularizer. The value of $\bar{p}$ in the regularization term is calculated as presented in Equation \ref{msn:reg}:

\begin{equation}
       \label{msn:reg}
       \bar{p} = \frac{1}{MB} \sum^{B}_{i=1}\sum^{M}_{m=1} p_{i,m}
\end{equation}

 Concerning masking, we use a masking ratio of $0.15$ for MSN and $0.6$ for MAE. We use exponential moving average with a momentum value of $0.996$ to update the \textit{target} encoder in MSN and the \textit{teacher} encoder in DINO, along with their respective projection heads. For focal view generation in MSN, we crop the image to $96 \times 96$. We set $M=11$ anchor views of an image and generate 1 random mask and 10 focal masks. 
 We also use ImageNet \cite{deng2009imagenet} statistics to normalize the input CXR scans during pretraining.

\subsection{Model Pretraining}

 We follow the same hyperparameter settings as in the original MSN \citep{assran2022masked}, DINO \citep{caron2021emerging}, and MAE \citep{he2022masked}. We use the AdamW optimizer \citep{loshchilov2017decoupled} for MSN and the other self-supervised baselines with a batch size of $64$. To select the learning rate, we conduct several experiments using different combinations of learning rate and weight decay values. We empirically found that the best learning rate and weight decay values for MSN are  $1e-4$ and $1e-3$, respectively, while the best values for DINO and MAE are $1e-5$ and $1e-4$, respectively. For all pertaining experiments, we employ the cosine annealing method as a learning rate scheduler. We pretrain each model for $100$ epochs with early stopping, using a patience of $5$ epochs and a minimum delta of $1e-5$ for the training loss.

 To incorporate EHR data in our proposal, we implement both $f_{ehr}$ and $g$ as linear layers (see Figure \ref{fig:method}). We fix the output dimension of $f_{ehr}$ as $128$ and match the output dimension of $g$ to the hidden size $D$ according to the backbone used (see Table \ref{tab:models}). %For the fusion operator $\oplus$, we utilize simple concatenation over the first dimension of both $v_{ehr}$ and $v_{cxr}$.
 Table \ref{tab:models} provides an overview of the size and architecture of the Vision Transformer architectures used.

 \begin{table}[h]
    \centering
    \caption{\small\textbf{Summary of Vision Transformer (ViT) model variants.} Descriptive summary of the ViT variants employed in our setup and the original Vision Transformer architecture, ViT-Base.}
    \resizebox{0.6\linewidth}{!}
    {\begin{tabular}{l l l l l l l l} 
\hline
 \textbf{Model} & \textbf{Layers} &\textbf{Hidden size $D$} & \textbf{MLP size} & \textbf{Heads} & \textbf{Parameters} \\ 
\hline
 ViT-B & 12 & 768 & 3072 & 6 & 86M \\
 ViT-S & 12 & 384 & 1536 & 6 & 21.7M \\
 ViT-T & 12 & 192 & 768 & 6 & 5.5M \\
\hline
\end{tabular}}
\label{tab:models}

\end{table}

\subsection{Downstream Classification}
\label{section:ds-setup}
 We use the Adam optimizer \citep{kingma2014adam} with a batch size of $64$ for all downstream evaluation experiments. For each method evaluation, we perform three experiments with learning rates in [$1e-3$,$5e-4$,$1e-4$] following \citep{azizi2021big}, each with cosine annealing learning rate scheduler. In this setup, we apply early stopping with a patience of 5 epochs if the validation AUROC does not improve by $1e-4$. Additionally, we conduct two more experiments with learning rates in [$5e-4$, $1e-4$] and use the Reduce on Plateau learning rate scheduler without early stopping. We perform this setup once for linear evaluation and replicate it for fine-tuning experiments, resulting in $10$ experiments for each model. We select the best-performing model on the validation set for test set evaluation. The maximum training epochs for all downstream evaluation tasks are set to 50 epochs. For augmentations, we apply random horizontal flip, random affine, and center crop during training, and only center crop during inference. We use ImageNet \citep{deng2009imagenet} statistics to normalize the CXR images during both training and inference, following the same settings as during pretraining.

 For low-data regime fine-tuning, we conduct five experiments on five randomly selected subsets of the training set, corresponding to specified percentages ($1\%$, $5\%$, and $10\%$), and evaluate the model on the full test set. To run these experiments, we use the same settings as the best-performing fine-tuning setup on the full training set. The best performing run is selected from the five runs based on the AUROC metric. 

% \subsection{External validation setup}
% \label{section:ext-val}
% For the CheXpert dataset \cite{irvin2019chexpert}, only the test set is utilized as the labels exactly match those in MIMIC-CXR \cite{johnson2019mimic}, and the training data is ignored. For evaluation, we load the weights of the best-performing models from the internal validation of each proposed strategy and assess the results on the CheXpert dataset against MSN. For the NIH-14 dataset \cite{wang2017chestx}, we conduct new experiments following the same setup mentioned in Section \ref{section:ds-setup} due to differences in labels between the two datasets, and the best-performing model is selected accordingly. In both datasets, evaluation is performed under linear evaluation and fine-tuning protocols.

% For the supervised baselines, we follow the procedure We apply data augmentations during training, including random horizontal flip, random rotation, and center crop We used Adam optimizer with default parameters and a learning rate of 0.0005. We decay the lr by 2 every time the loss does not improve for 3 epoch.  

%\subsection{General training setup}
All pretraining experiments are conducted using a single Nvidia A100 GPU, and downstream evaluation is performed using a single Nvidia V100 GPU. Pretraining takes approximately 20-30 hours for $100$ epochs, while downstream evaluation requires 2-5 hours for $50$ epochs, depending on the backbone used. We also enable automatic mixed-precision \citep{micikevicius2017mixed} in all experiments to speed up training and reduce memory requirements.

% \setcounter{table}{0}
% \renewcommand{\thetable}{B\arabic{table}}
% \renewcommand{\thefigure}{B\arabic{figure}}

% \newpage
% \section{Masked Siamese Networks (MSN)}
% \label{appendix:msn}

%\newpage

\setcounter{table}{0}
\setcounter{figure}{0}
\renewcommand{\thetable}{B\arabic{table}}
\renewcommand{\thefigure}{B\arabic{figure}}
% \section{Datasets}

\newpage
\section{EHR Features}
\label{appendix:datasets}

\begin{figure}[ht]
    \centering
    \includegraphics[width=0.8\linewidth]{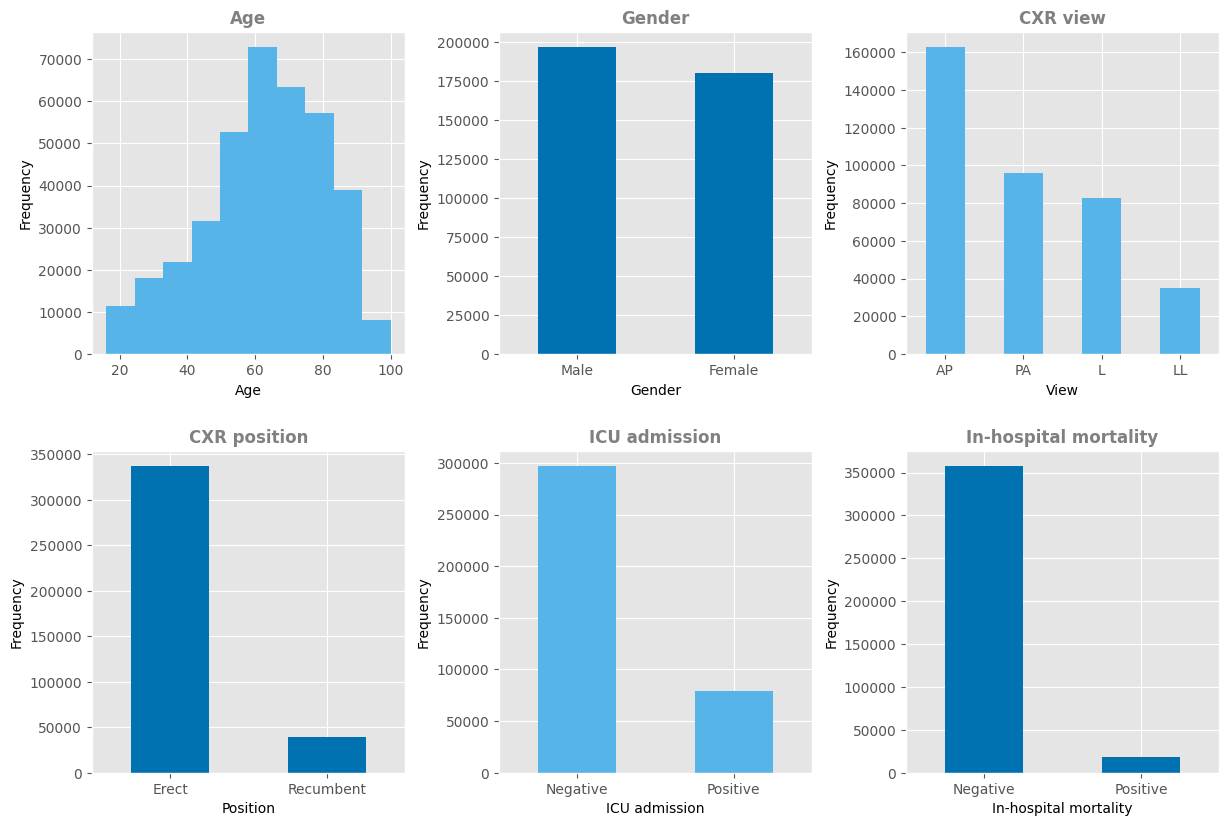}
    \caption{\textbf{Distribution of EHR features.} Absolute frequency distributions of the EHR features employed in our experimental setup. Note that the Y-axis in all plots provides the absolute frequency value, but the scale differs.}
    \label{fig:distributions}
\end{figure}

Figure \ref{fig:distributions} provides the frequency distributions for the EHR features considered in our work during the self-supervised pretraining stage. 

\setcounter{table}{0}
\renewcommand{\thetable}{B\arabic{table}}

\newpage

% \newpage

\setcounter{table}{0}
\setcounter{figure}{0}
\renewcommand{\thetable}{C\arabic{table}}
\renewcommand{\thefigure}{C\arabic{figure}}

\section{Additional Results}
\label{section:add-results}
In this section, we present additional results from our experimental setting. These results complement or extend the ones provided in the main text, including additional baselines or architectures (i.e., ViT-T).

\subsection{Linear Evaluation}
\label{app:results_linear_evaluation}

Table \ref{tab:linear-eval-additional} provides performance results for linear evaluation of self-supervised methods using the ViT-T architecture as backbone model. 

\begin{table}[!ht]
    \centering
    \caption{\small\textbf{Performance results for linear evaluation of self-supervised methods using ViT-T as backbone model}. We summarize the AUROC and AUPRC performance on the test set for the self-supervised methods as well as the 95\% confidence intervals. The best results are shown in blue. We use {\color{green}$\uparrow$} and {\color{red}$\downarrow$} to indicate whether the performance of a given model is 0.5-1.5\% better or worse than the reference model (i.e., vanilla MSN).    {\color{green}$\uparrow\uparrow$} and {\color{red}$\downarrow\downarrow$} indicate whether the difference is 1.5\% better or worse than the reference model, respectively. $-$ indicates that the performance difference is less than 0.5\% compared to the reference model.}
    \resizebox{0.5\linewidth}{!}
    {\begin{tabular}{l l l} 
    \hline
    %\textbf{Dataset} & \multicolumn{2}{c}{\textbf{MIMIC-CXR}} \\ \midrule
    \textbf{Pretraining} &\textbf{AUROC (CI)} & \textbf{AUPRC (CI)} \\
    \hline
    ImageNet &  0.684 (0.680, 0.688) \color{red}$\downarrow\downarrow$ & 0.251 (0.249, 0.254) \color{red}$\downarrow\downarrow$ \\ 
   % & &   &  \\ 
    DINO &  0.694 (0.691, 0.698) \color{red}$\downarrow$ & 0.262 (0.259, 0.264) \color{red}$\downarrow$ \\ 
 % % & &  &   \\ \hline
    MAE &  0.633 (0.629, 0.637) \color{red}$\downarrow\downarrow$ & 0.211 (0.209, 0.213) \color{red}$\downarrow\downarrow$ \\ 
    MSN &  0.708 (0.704, 0.711)  & 0.272 (0.270, 0.275)  \\ \hline
  
    MSN$+ x_{sex}$  &  0.728$^*$ (0.724, 0.731) \color{green}$\uparrow\uparrow$ &  0.290$^*$ (0.288, 0.293) \color{green}$\uparrow\uparrow$\\   
    MSN$+ x_{age}$  &  0.728$^*$ (0.724, 0.732) \color{green}$\uparrow\uparrow$ &  0.289$^*$ (0.287, 0.292) \color{green}$\uparrow\uparrow$\\
    MSN+$x_{view}$  &  0.726$^*$ (0.722, 0.730) \color{green}$\uparrow\uparrow$  & 0.288$^*$ (0.286, 0.291) \color{green}$\uparrow\uparrow$ \\
    MSN+$x_{pos}$  & \color{blue}{0.729$^*$ (0.726, 0.733)} \color{green}$\uparrow\uparrow$ & \color{blue}{0.292$^*$ (0.290, 0.295)} \color{green}$\uparrow\uparrow$ \\
    MSN+$x_{mort}$  &  0.727$^*$ (0.723, 0.730)	\color{green}$\uparrow\uparrow$ & 0.291$^*$ (0.288, 0.294) \color{green}$\uparrow\uparrow$ \\
    MSN+$x_{icu}$  &  0.726$^*$ (0.722, 0.729) \color{green}$\uparrow\uparrow$ & 0.289$^*$ (0.287, 0.292) \color{green}$\uparrow\uparrow$ \\
    MSN+$x_{D}$ &  0.727$^*$ (0.723, 0.731) \color{green}$\uparrow\uparrow$ & 0.290$^*$ (0.287, 0.293) \color{green}$\uparrow\uparrow$ \\ 
    MSN+$x_{SM}$  & 0.727$^*$ (0.723, 0.730) \color{green}$\uparrow\uparrow$ & 0.290$^*$ (0.288, 0.293) \color{green}$\uparrow\uparrow$ \\
    MSN+$x_{SI}$  &  0.725$^*$ (0.721, 0.728)	\color{green}$\uparrow\uparrow$ & 0.289$^*$ (0.286, 0.292) \color{green}$\uparrow\uparrow$ \\
    MSN+$x_{D+SM}$  & 0.724$^*$ (0.720, 0.727)	 \color{green}$\uparrow\uparrow$ & 0.285$^*$ (0.283, 0.288) \color{green}$\uparrow$ \\
    MSN+$x_{D+SI}$  & 0.723$^*$ (0.720, 0.727)  \color{green}$\uparrow\uparrow$ & 0.287$^*$ (0.285, 0.291) \color{green}$\uparrow\uparrow$\\
    MSN+$x_{SM+SI}$  & 0.726$^*$ (0.723, 0.729) \color{green}$\uparrow\uparrow$ & 0.288$^*$ (0.286, 0.291) \color{green}$\uparrow\uparrow$ \\
    MSN+$x_{D+SM+SI}$  &  0.724$^*$ (0.720, 0.727) \color{green}$\uparrow\uparrow$ & 0.287$^*$ (0.284, 0.289) \color{green}$\uparrow\uparrow$\\
    \hline
    \multicolumn{3}{l}{{\color{black}\small $^*$ Statistical significance results with respect to vanilla MSN ($p < 0.001$)}} \\
    \end{tabular}}
    \label{tab:linear-eval-additional}
    %\vspace{-4mm}
    \end{table}

% \subsection{t-SNE Results.}

% \label{appendix_t-sne_results}

% Figure \ref{fig:t-sne} visualizes and compares the quality of the embeddings obtained with vanilla MSN (baseline), MSN$+ x_{sex}$, MSN$+ x_{D}$, and MSN$+ x_{D,SM,SI}$. For the ease of interpretability, we apply t-SNE to a subset of samples with a single label in the training set (n=68000), considering the challenge of visualizing multi-label samples in the latent space, which is not the main objective of this work.

% \begin{figure}[!ht]
%     \centering
%     \includegraphics[scale=0.4,width=\linewidth]{MIDL-v/images/tsne-results.jpeg}
%     \caption{\textbf{Embedding visualization via t-SNE plots.} (a): Vanilla MSN (b): MSN$+ x_{sex}$ (c):  MSN$+ x_{D}$ (e): MSN$+ x_{D+SM+SI}$}
%     \label{fig:t-sne}
% \end{figure}

\subsection{Supervised learning}

Table \ref{tab:supervised_learning} reports performance results of reference architectures trained using supervised learning for comparison with self-supervised methods. 

\begin{table}[!ht]
    \centering
    \caption{\small\textbf{Supervised learning models}. Reference models trained using supervised learning.}
    \resizebox{0.6\linewidth}{!}
    {\begin{tabular}{l l l l} 
    \hline
    \textbf{Backbone} & \textbf{Initialization} & \textbf{AUROC (CI)}  &\textbf{AUPRC (CI)} \\ 
    \hline
  ViT-T & Random &  0.762 (0.758, 0.765) & 0.329 (0.326, 0.332)\\ 
  ViT-T & ImageNet &  0.783 (0.780, 0.787)& 0.356 (0.353, 0.360)\\ \hline
  ViT-S & Random & 0.763 (0.759, 0.766) & 0.330 (0.328, 0.334) \\
  ViT-S & ImageNet & 0.782  (0.778, 0.785) & 0.353 (0.350, 0.358) \\ 
  \hline
\end{tabular}}
\label{tab:supervised_learning}
%\vspace{-4mm}
\end{table}

\subsection{Fine-tuning}

% \subsection{Fine-tuning results}

% Table \ref{tab:ft-small} reports the performance results for fine-tuning experiments under low-data regimes.
% % As in the linear evaluation, we include different pre-initialization strategies of the supervised baselines (ImageNet and random) and combinations of EHR additions to the vanilla MSN architecture. 
% In general, we observe that our pretraining strategies yield performance that is, at worst, on par with the vanilla MSN when fine-tuning using the full training set. We attribute this performance to the absence of integrated EHR data during the fine-tuning stage, which was present during pretraining, as well as the weight changes that occur during fine-tuning. On the other hand, our proposed framework demonstrates improved downstream classification for end-to-end fine-tuned models in low-data regimes, particularly with 1\% of the training data and socio-demographic variables, as shown in Table \ref{tab:ft-small}. The performance remains on par with vanilla MSN as the percentage of training data increases. It is worth noting that our proposed strategies consistently improve performance compared to the DINO and MAE baselines under both low and high data regimes.

Table \ref{tab:ft-small} provides performance results for fine-tuning of self-supervised methods using the ViT-S architecture as backbone model. Table \ref{tab:ft-tiny} provides the results obtained using ViT-T architecture as backbone.

\begin{table}[!ht]
    \centering
    \caption{\small\textbf{Performance results for the fine-tuning of our proposed methods using ViT-S as backbone and under low-data regime.} We summarize AUROC and AUPRC results on the test set including 95\% confidence intervals. The best results are shown in blue.We use {\color{green}$\uparrow$} and {\color{red}$\downarrow$} to indicate whether the performance of a given model is 0.5-1.5\% better or worse than the reference model (i.e., vanilla MSN).    {\color{green}$\uparrow\uparrow$} and {\color{red}$\downarrow\downarrow$} indicate whether the difference is 1.5\% better or worse than the reference model, respectively. $-$ indicates that the performance difference is less than 0.5\% compared to the reference model.}
     \resizebox{\linewidth}{!}
    {\begin{tabular}{l l l l l l l } 
\hline

\multirow{3}{*}{\textbf{Pretraining}} & \multicolumn{6}{c}{\textbf{Training data proportion}} \\ \cline{2-7}
& \multicolumn{2}{c}{\textbf{1\%}} & \multicolumn{2}{c}{\textbf{5\%}} & \multicolumn{2}{c}{\textbf{10\%}} \\ \cline{2-7} %&  \multicolumn{2}{c}{\textbf{100\%}} \\ \cline{2-9}
& \textbf{AUROC (CI)} & \textbf{AUPRC (CI)} & \textbf{AUROC (CI)} & \textbf{AUPRC (CI)} & \textbf{AUROC (CI)} & \textbf{AUPRC (CI)} \\ \hline %& \textbf{AUROC (CI)} & \textbf{AUPRC (CI)} \\ \hline
  
DINO & 0.690 (0.687, 0.694) \color{red}$\downarrow\downarrow$ & 0.269 (0.267, 0.271) \color{red}$\downarrow\downarrow$ & 0.730 (0.727, 0.734)\color{red}$\downarrow\downarrow$&0.302 (0.300, 0.305)\color{red}$\downarrow$ & 0.737 (0.733, 0.740)\color{red}$\downarrow\downarrow$& 0.305 (0.302, 0.308)\color{red}$\downarrow\downarrow$ \\ % &  0.769 (0.766, 0.773) & 0.341 (0.338, 0.344) \\
MAE  & 0.659 (0.655, 0.664) \color{red}$\downarrow\downarrow$ & 0.241 (0.240, 0.244) \color{red}$\downarrow\downarrow$ &0.702 (0.699, 0.706)\color{red}$\downarrow\downarrow$ &0.274 (0.272, 0.277)\color{red}$\downarrow\downarrow$ & 0.727 (0.723, 0.731)\color{red}$\downarrow\downarrow$& 0.297 (0.294, 0.300)\color{red}$\downarrow\downarrow$ \\ %& 0.765 (0.762, 0.769) & 0.332 (0.329, 0.336) \\
MSN & 0.707 (0.703, 0.711)  & 0.286 (0.284, 0.289)  & 0.746 (0.743, 0.75)  & 0.316 (0.313, 0.319)  & 0.760 (0.756, 0.763)  & 0.329 (0.327, 0.333)  \\ \hline  %&  0.793 (0.789, 0.796) & 0.367 (0.363, 0.371)\\ \hline

MSN$+ x_{sex}$ & 0.713 (0.709, 0.717) \color{green}$\uparrow$ & 0.289 (0.287, 0.292) $-$  & \color{blue}{0.751 (0.747, 0.754)} \color{green}$\uparrow$  & 0.315 (0.313, 0.319) $-$ & 0.761 (0.758, 0.765) $-$ & 0.329 (0.327, 0.333) $-$ \\ %&   0.788 (0.785, 0.791) &  0.358 (0.354, 0.362)\\   

MSN$+ x_{age}$ & \color{blue}{0.714 (0.710, 0.718)} \color{green}$\uparrow$ & \color{blue}{0.290 (0.288, 0.293)} $-$ & \color{blue}{0.751 (0.748, 0.755)} \color{green}$\uparrow$ 	& \color{blue}{0.319 (0.316, 0.322)} $-$ & 0.760 (0.756, 0.763) $-$ &	0.326 (0.324, 0.330) $-$ \\ %&  0.789 (0.786, 0.792)	 &  0.360 (0.356, 0.364)\\

MSN$+ x_{view}$ & 0.708 (0.704, 0.711) $-$ & 0.282 (0.280, 0.285) $-$ & 0.748 (0.744, 0.751) $-$	& 0.317 (0.315, 0.320) $-$ &0.761 (0.757, 0.764) $-$ &0.329 (0.327, 0.333) $-$ \\ %& 0.789 (0.786, 0.792)	 &  0.359 (0.356, 0.364)\\

MSN$+ x_{pos}$ & 0.713 (0.710, 0.717) \color{green}$\uparrow$  & \color{blue}{0.290 (0.288, 0.292)} $-$  & 	0.749 (0.745, 0.752) $-$ & 0.317 (0.315, 0.320) $-$ &0.761 (0.758, 0.764) $-$  &0.329 (0.326, 0.333) $-$ \\ %& 0.788 (0.785, 0.791) &  0.360 (0.357, 0.364)\\
MSN$+ x_{mort}$ & 0.711 (0.707, 0.715) $-$ & 0.282 (0.279, 0.284) $-$  & 0.748 (0.744, 0.751) $-$	& 0.317 (0.314, 0.320) $-$ & 0.761 (0.757, 0.764) $-$ & 0.326 (0.323, 0.329) $-$ \\ %& 0.790 (0.787, 0.793)	  &  0.362 (0.359, 0.367)\\

MSN$+ x_{icu}$ & 0.706 (0.702, 0.710) $-$ & 0.284 (0.282, 0.287) $-$  & 0.747 (0.744, 0.751) $-$	& 0.314 (0.311, 0.317) $-$ & 0.758 (0.755, 0.762) $-$ & 0.324 (0.321, 0.327) $-$ \\ %& 0.788 (0.785, 0.791)	&  0.359 (0.356, 0.363)\\

MSN$+ x_{D}$ & 0.710 (0.706, 0.713) $-$ & \color{blue}{0.290 (0.288, 0.293)} $-$ & 	0.749 (0.746, 0.752) $-$ & 0.315 (0.313, 0.318) $-$ & \color{blue}{0.763 (0.760, 0.766)} $-$ & \color{blue}{0.330 (0.327, 0.333)} $-$ \\ %&  0.790 (0.786, 0.792) & 0.361 (0.358, 0.365) \\  	
MSN$+ x_{SM}$ & 0.710 (0.706, 0.714) $-$ & 0.287 (0.285, 0.290) $-$ &  0.748 (0.744, 0.751) $-$ & 0.315 (0.313, 0.319) $-$ & 0.758 (0.755, 0.762) $-$ & 0.327 (0.324, 0.330) $-$ \\ %& 0.787 (0.784, 0.79)	  &  0.358 (0.355, 0.363) \\
MSN$+ x_{SI}$ & 0.708 (0.704, 0.712) $-$ & 0.286 (0.284, 0.289) $-$ & 0.749 (0.746, 0.753) $-$ & 0.315 (0.313, 0.318) $-$ & 0.762 (0.759, 0.766) $-$ &	0.327 (0.325, 0.331) $-$ \\ %&  0.788 (0.785, 0.791)	 &  0.359 (0.356, 0.363)\\
MSN$+ x_{D+SM}$ & 0.704 (0.701, 0.708) $-$ & 0.287 (0.285, 0.290) $-$ & 0.746 (0.742, 0.749) $-$ & 0.314 (0.312, 0.317) $-$ & 0.758 (0.754, 0.761) $-$ &	0.325 (0.323, 0.329) $-$ \\ % &  0.787 (0.784, 0.789)	 &  0.357 (0.354, 0.361)\\
MSN$+ x_{D+SI}$ & 0.703 (0.699, 0.706) $-$ & 0.286 (0.284, 0.288) $-$ & 0.745 (0.742, 0.749) $-$ & 0.315 (0.313, 0.318) $-$ & 0.762 (0.759, 0.765) $-$ &	0.328 (0.325, 0.331) $-$ \\ % &  0.787 (0.784, 0.79)	&  0.360 (0.356, 0.364) \\
MSN$+ x_{SM+SI}$ & 0.708 (0.704, 0.711) $-$ & 0.282 (0.280, 0.284) $-$  & 	0.746 (0.743, 0.750) $-$ & 0.315 (0.313, 0.319) $-$ & 0.760 (0.757, 0.763) $-$  &0.326 (0.323, 0.329) $-$ \\ %&  0.789 (0.786, 0.792) &  0.361 (0.357, 0.365) \\
MSN$+ x_{D+SM+SI}$ & 0.706 (0.702, 0.709) $-$  & 0.285 (0.283, 0.288) $-$ & 0.746 (0.743, 0.750)	$-$ & 0.314 (0.312, 0.318) $-$ & 0.756 (0.753, 0.759) $-$ & 0.326 (0.323, 0.330) $-$ \\ % & 0.787 (0.784, 0.79)	 &  0.357 (0.354, 0.362)\\

\hline
\end{tabular}}
\label{tab:ft-small}
\end{table}

\begin{table}[!ht]
    \centering
    \caption{\small\textbf{Performance results for the fine-tuning of our proposed methods using ViT-T as backbone and under low-data regime.} We summarize AUROC and AUPRC results on the test set including 95\% confidence intervals. The best results are shown in blue. We use {\color{green}$\uparrow$} and {\color{red}$\downarrow$} to indicate whether the performance of a given model is 0.5-1.5\% better or worse than the reference model (i.e., vanilla MSN).    {\color{green}$\uparrow\uparrow$} and {\color{red}$\downarrow\downarrow$} indicate whether the difference is 1.5\% better or worse than the reference model, respectively. $-$indicates that the performance difference is less than 0.5\% compared to the reference model.}
     \resizebox{\linewidth}{!}
    {\begin{tabular}{l l l l l l l } 
\hline

\multirow{3}{*}{\textbf{Pretraining}} & \multicolumn{6}{c}{\textbf{Training data proportion}} \\ \cline{2-7}
& \multicolumn{2}{c}{\textbf{1\%}} & \multicolumn{2}{c}{\textbf{5\%}} & \multicolumn{2}{c}{\textbf{10\%}} \\ \cline{2-7} %\multicolumn{2}{c|}{\textbf{remove}} & \multicolumn{2}{c}{\textbf{100\%}} \\ \cline{2-7}
& \textbf{AUROC (CI)} & \textbf{AUPRC (CI)} & \textbf{AUROC (CI)} & \textbf{AUPRC (CI)} & \textbf{AUROC (CI)} & \textbf{AUPRC (CI)} \\ \hline %& \textbf{AUROC (CI)} & \textbf{AUPRC (CI)} \\ \hline

DINO & 0.685 (0.682, 0.689) \color{red}$\downarrow$ & 0.266 (0.264, 0.268) \color{red}$\downarrow$ & 0.727 (0.723, 0.730) \color{red}$\downarrow\downarrow$ & 0.297 (0.295, 0.300) \color{red}$\downarrow$ & 0.741 (0.737, 0.745) \color{red}$\downarrow$ &0.310 (0.308, 0.313) \color{red}$\downarrow$ \\ %& & & 0.771 (0.768, 0.775) & 0.342 (0.339, 0.346) \\ 
MAE  &0.640 (0.636, 0.644) \color{red}$\downarrow\downarrow$ & 0.228 (0.227, 0.231) \color{red}$\downarrow\downarrow$ & 0.700 (0.697, 0.704) \color{red}$\downarrow\downarrow$ & 0.271 (0.269, 0.274) \color{red}$\downarrow\downarrow$ & 0.721 (0.717, 0.725) \color{red}$\downarrow\downarrow$ & 0.289 (0.286, 0.292) \color{red}$\downarrow\downarrow$ \\ %& & & 0.763 (0.759, 0.766) & 0.330 (0.327, 0.333)\\ 
MSN &0.694 (0.69, 0.698)  &0.276 (0.274, 0.279)  &  \color{blue}{0.742 (0.739, 0.746)} & 0.311 (0.309, 0.314) & \color{blue}{0.754 (0.751, 0.758)}   & \color{blue}{0.322 (0.320, 0.326)} \\ \hline %& & & 0.789 (0.786, 0.792) & 0.361 (0.357, 0.366) \\ \hline

MSN$+ x_{sex}$  & 0.700 (0.696, 0.704) \color{green}$\uparrow$ & 0.280 (0.278, 0.283) $-$ & 0.739 (0.735, 0.742) $-$ & 0.310 (0.308, 0.313) $-$ & \color{blue}{0.754 (0.750, 0.758)} $-$ & 0.320 (0.318, 0.324) $-$ \\ % & & &  0.783 (0.781, 0.787)&  0.352 (0.349, 0.356)  \\

MSN$+ x_{age}$  &0.697 (0.693, 0.701) $-$ &0.278 (0.276, 0.281) $-$ & 0.740 (0.737, 0.744) $-$ & \color{blue}{0.312 (0.310, 0.315)} $-$ & 0.753 (0.750, 0.756) $-$ & 0.321 (0.318, 0.324) $-$ \\ %& & & 0.785 (0.781, 0.788)	  &  0.356 (0.353, 0.36)\\

MSN$+ x_{view}$  & 0.698 (0.693, 0.701) $-$ & 0.280 (0.278, 0.283) $-$ & \color{blue}{0.742 (0.738, 0.746)} $-$ & 0.310 (0.307, 0.313) $-$ & 0.752 (0.748, 0.755) $-$ & 0.320 (0.317, 0.323) $-$ \\ %& & & 0.783 (0.78, 0.786)	&  0.354 (0.350, 0.358)\\

MSN$+ x_{pos}$  & \color{blue}{0.702 (0.698, 0.705)} \color{green}$\uparrow$ & 0.279 (0.277, 0.281) $-$ &0.739 (0.734, 0.742) $-$  & 0.308 (0.306, 0.312) $-$ & \color{blue}{0.754 (0.750, 0.757)} $-$ & 0.319 (0.316, 0.322) $-$ \\ %& & &  0.782 (0.779, 0.785)	 &  0.354 (0.35, 0.358) \\

MSN$+ x_{mort}$  & 0.698 (0.694, 0.702) $-$ & \color{blue}{0.283 (0.280, 0.285)} \color{green}$\uparrow$ & 0.738 (0.735, 0.741) $-$ & 0.307 (0.305, 0.311) $-$ & 0.752 (0.748, 0.755) $-$ & 0.318 (0.316, 0.321) $-$ \\ %& & & 0.783 (0.780, 0.786)	 &  0.354 (0.351, 0.358)\\

MSN$+ x_{icu}$  & 0.700 (0.696, 0.704) \color{green}$\uparrow$ & 0.278 (0.276, 0.281) $-$ &0.739 (0.736, 0.743) $-$ & 0.308 (0.305, 0.311) $-$ & 0.751 (0.748, 0.754) $-$ & 0.319 (0.316, 0.322) $-$ \\ %& & & 0.785 (0.782, 0.788)	 &  0.354 (0.35, 0.358)\\

MSN$+ x_{D}$ & 0.698 (0.694, 0.701) $-$ &0.273 (0.271, 0.276) $-$  & 0.738 (0.734, 0.742) $-$ & 0.309 (0.306, 0.311) $-$ & 0.752 (0.748, 0.755) $-$ & 0.318 (0.316, 0.321) $-$ \\ %& & & 0.788 (0.784, 0.791) & 0.357 (0.354, 0.361) \\ 

MSN$+ x_{SM}$  & 0.699 (0.695, 0.703) \color{green}$\uparrow$ & 0.278 (0.276, 0.281) $-$ & 0.740 (0.736, 0.743) $-$ & 0.308 (0.306, 0.311) $-$ & \color{blue}{0.754 (0.75, 0.757)} $-$ & \color{blue}{0.322 (0.320, 0.326)} $-$ \\ % & & & 0.782 (0.779, 0.785)	 &  0.353 (0.349, 0.357)\\

MSN$+ x_{SI}$  & \color{blue}{0.702 (0.698, 0.706)} \color{green}$\uparrow$ & 0.282 (0.280, 0.285) \color{green}$\uparrow$ & 0.739 (0.736, 0.743) $-$ & 0.309 (0.306, 0.311) $-$ & 0.750 (0.746, 0.754) $-$ & 0.317 (0.315, 0.320) \color{red}$\downarrow$ \\ %& & & 0.783 (0.779, 0.786)	 &  0.352 (0.349, 0.356) \\

MSN$+ x_{D+SM}$  & 0.697 (0.693, 0.701) $-$ & 0.277 (0.275, 0.280) $-$ &0.735 (0.731, 0.738) \color{red}$\downarrow$  & 0.307 (0.304, 0.310) $-$ & 0.749 (0.745, 0.752) \color{red}$\downarrow$ & 0.317 (0.315, 0.320) \color{red}$\downarrow$ \\ %& & & 0.78 (0.777, 0.783)	 &  0.351 (0.348, 0.355)\\

MSN$+ x_{D+SI}$  &0.697 (0.694, 0.701) $-$ & 0.278 (0.276, 0.281) $-$ & 0.740 (0.737, 0.744) $-$ &0.307 (0.304, 0.310) $-$ & 0.753 (0.750, 0.757) $-$ & 0.321 (0.318, 0.324) $-$ \\ %& & & 0.784 (0.781, 0.787)	  &  0.353 (0.35, 0.357)\\

MSN$+ x_{SM+SI}$  & 0.699 (0.694, 0.702) \color{green}$\uparrow$ & 0.276 (0.273, 0.278) $-$ & 0.739 (0.735, 0.742) $-$ & 0.306 (0.304, 0.310) \color{red}$\downarrow$ & 0.750 (0.746, 0.753) $-$ & 0.316 (0.313, 0.319) \color{red}$\downarrow$ \\ % & & & 0.783 (0.78, 0.786) &  0.352 (0.349, 0.356) \\

MSN$+ x_{D+SM+SI}$ &0.701 (0.698, 0.705) \color{green}$\uparrow$ & 0.280 (0.278, 0.283) $-$ & 0.736 (0.733, 0.740) \color{red}$\downarrow$ & 0.305 (0.303, 0.308) \color{red}$\downarrow$ &0.752 (0.749, 0.756) $-$  & 0.321 (0.319, 0.325) $-$ \\ %& & & 0.780 (0.777, 0.783) &  0.348 (0.345, 0.352)\\ 

\hline
\end{tabular}}
\label{tab:ft-tiny}
\end{table}

\subsection{External Validation}

Table \ref{tab_appendix:ext-val-t-s} reports external validation results on downstream classification for fine tuning of ViT-S self-supervised models on the ChexPert and NIH-14 datasets.

\begin{table}[!ht]
    \centering
    \caption{\small{\textbf{Performance results for external validation of our proposed methods using ViT-S backbone under fine-tuning self-supervised.} We summarize AUROC and AUPRC results on CheXpert and NIH-14 test sets including 95\% confidence intervals. The best results are shown in blue. We use {\color{green}$\uparrow$} and {\color{red}$\downarrow$} to indicate whether the performance of a given model is 0.5-1.5\% better or worse than the reference model (i.e., vanilla MSN).    {\color{green}$\uparrow\uparrow$} and {\color{red}$\downarrow\downarrow$} indicate whether the difference is 1.5\% better or worse than the reference model, respectively. $-$ indicates that the performance difference is less than 0.5\% compared to the reference model.}}
    \resizebox{0.8\linewidth}{!}
    {
    \begin{tabular}{l l l  l l }
        \hline
         &\multicolumn{2}{c}{\textbf{CheXpert}} & \multicolumn{2}{c}{\textbf{NIH-14}} \\\hline
         & \textbf{AUROC (CI)} & \textbf{AUPRC (CI)}&\textbf{AUROC (CI)} & \textbf{AUPRC (CI)} \\\hline
         MSN & 0.844 (0.827, 0.862) & 0.505 (0.483, 0.536) & \color{blue}{0.783 (0.780, 0.787)}  &0.346 (0.339, 0.359) \\\hline
         
         MSN$+ x_{sex}$ & 0.850 (0.832, 0.867) \color{green}$\uparrow$ & 0.505 (0.480, 0.541) $-$ & 0.782 (0.778, 0.787) $-$ & \color{blue}{0.350 (0.341, 0.361)} $-$ \\
         
         MSN$+ x_{age}$ & 0.847 (0.830, 0.863) $-$ & 0.491 (0.471, 0.522) \color{red}$\downarrow$ & \color{blue}{0.783 (0.779, 0.787)} $-$ & \color{blue}{0.350 (0.342, 0.360)} $-$ \\
         
         MSN$+ x_{view}$ & 0.837 (0.819, 0.855) \color{red}$\downarrow$ & 0.487 (0.470, 0.514) \color{red}$\downarrow\downarrow$ & 0.777 (0.772, 0.781) \color{red}$\downarrow$  & 0.338 (0.330, 0.349) \color{red}$\downarrow$ \\
         
         MSN$+ x_{pos}$  & 0.836 (0.819, 0.852) \color{red}$\downarrow$ & 0.498 (0.476, 0.529) \color{red}$\downarrow$ & 0.782 (0.778, 0.785) $-$ & 0.343 (0.337, 0.354) $-$ \\
         
         MSN$+ x_{mort}$ & 0.841 (0.823, 0.859) $-$ & 0.521 (0.494, 0.559) \color{green}$\uparrow\uparrow$ & 0.782 (0.776, 0.786) $-$ & 0.343 (0.336, 0.356) $-$ \\
         
         MSN$+ x_{icu}$  &  0.847 (0.829, 0.867) $-$ & \color{blue}{0.510 (0.487, 0.550)} \color{green}$\uparrow$ &  0.781 (0.778, 0.786) $-$ & 0.343 (0.337, 0.354) $-$ \\
         
         MSN$+ x_{D}$  & 0.830 (0.815, 0.846) \color{red}$\downarrow$ & 0.477 (0.458, 0.507) \color{red}$\downarrow\downarrow$ & \color{blue}{0.783 (0.777, 0.787)} $-$ & 0.347 (0.339, 0.358) $-$ \\ 
         
         MSN$+ x_{SM}$  & 0.828 (0.807, 0.848) \color{red}$\downarrow\downarrow$ & 0.483 (0.464, 0.516) \color{red}$\downarrow\downarrow$ & 0.780 (0.776, 0.784) $-$ &0.343 (0.335, 0.353) $-$ \\
         
         MSN$+ x_{SI}$  & 0.849 (0.830, 0.866) \color{green}$\uparrow$ &0.492 (0.474, 0.520) \color{red}$\downarrow$ & 0.780 (0.776, 0.784) $-$ & 0.347 (0.340, 0.357) $-$ \\
         
         MSN$+ x_{D+SM}$ & 0.850 (0.830, 0.869) \color{green}$\uparrow$ & \color{blue}{0.510 (0.487, 0.546)} \color{green}$\uparrow$ & 0.775 (0.771, 0.780) \color{red}$\downarrow$ & 0.338 (0.331, 0.349) \color{red}$\downarrow$ \\
         
         MSN$+ x_{D+SI}$ & 0.827 (0.807, 0.845) \color{red}$\downarrow\downarrow$ & 0.480 (0.462, 0.511) \color{red}$\downarrow\downarrow$ &  0.776 (0.772, 0.781) \color{red}$\downarrow$ &0.342 (0.335, 0.354) $-$ \\
         
         MSN$+ x_{SM+D}$ & 0.834 (0.816, 0.852) \color{red}$\downarrow$ & 0.489 (0.470, 0.525) \color{red}$\downarrow\downarrow$ & 0.782 (0.778, 0.786) $-$ & 0.339 (0.333, 0.347) \color{red}$\downarrow$ \\
         
         MSN$+ x_{D+SM+SI}$  & \color{blue}{0.854 (0.835, 0.870)} \color{green}$\uparrow$ & 0.503 (0.481, 0.534) $-$ & 0.775 (0.769, 0.779) \color{red}$\downarrow$ & 0.338 (0.331, 0.351) \color{red}$\downarrow$ \\
         \hline
         
    \end{tabular}}
    \label{tab_appendix:ext-val-t-s}
\end{table}

Table \ref{tab:ext-val-t-1} report external validation results on downstream classification for linear evaluation of ViT-S self-supervised models on the ChexPert and NIH-14 datasets. Table \ref{tab:ext-val-t-2} provides the results for the fine-tuning of self-supervised models.

\begin{table}[!ht]
    \centering
    \caption{\small{\textbf{Performance results of external validation for linear evaluation of self-supervised methods using ViT-T as backbone model}. We summarize AUROC and AUPRC results on CheXpert and NIH-14 test sets including 95\% confidence intervals. The best results are shown in blue. We use {\color{green}$\uparrow$} and {\color{red}$\downarrow$} to indicate whether the performance of a given model is 0.5-1.5\% better or worse than the reference model (i.e., vanilla MSN).    {\color{green}$\uparrow\uparrow$} and {\color{red}$\downarrow\downarrow$} indicate whether the difference is 1.5\% better or worse than the reference model, respectively. $-$ indicates that the performance difference is less than 0.5\% compared to the reference model.}}
    \resizebox{0.8\linewidth}{!}
    {
    \begin{tabular}{l l l l l }
        \hline
         &\multicolumn{2}{c}{\textbf{CheXpert}} & \multicolumn{2}{c}{\textbf{NIH-14}} \\\hline
         &\textbf{AUROC (CI)} & \textbf{AUPRC (CI)} & \textbf{AUROC (CI)} & \textbf{AUPRC (CI)} \\\hline
         MSN & 0.740 (0.720, 0.763)  & 0.396 (0.382, 0.421)  & 0.676 (0.671, 0.68)  & 0.199 (0.197, 0.203)  \\\hline
         
         MSN$+ x_{sex}$  & 0.742 (0.721, 0.764) $-$ & 0.403 (0.389, 0.425) \color{green}$\uparrow$ &0.711$^*$ (0.707, 0.714) \color{green}$\uparrow\uparrow$&0.233$^*$ (0.230, 0.242) \color{green}$\uparrow\uparrow$ \\
         
         MSN$+ x_{age}$  & 0.767 (0.746, 0.789) \color{green}$\uparrow\uparrow$ & 0.390 (0.378, 0.410) \color{red}$\downarrow$ & 0.711$^*$ (0.707, 0.715) \color{green}$\uparrow\uparrow$& 0.232$^*$ (0.229, 0.238) \color{green}$\uparrow\uparrow$ \\
         
         MSN$+ x_{view}$ & 0.765 (0.742, 0.788) \color{green}$\uparrow\uparrow$ & \color{blue}{0.420$^*$  (0.405, 0.453)} \color{green}$\uparrow\uparrow$ & 0.711$^*$ (0.708, 0.715) \color{green}$\uparrow\uparrow$&0.236$^*$ (0.232, 0.245) \color{green}$\uparrow\uparrow$ \\
         
         MSN$+ x_{pos}$  & 0.749 (0.728, 0.770) \color{green}$\uparrow$ & 0.392 (0.380, 0.411) $-$ & 0.710$^*$ (0.706, 0.713) \color{green}$\uparrow\uparrow$& 0.235$^*$ (0.232, 0.241) \color{green}$\uparrow\uparrow$\\
         
         MSN$+ x_{mort}$ & 0.750 (0.728, 0.772) \color{green}$\uparrow$ & 0.413 (0.400, 0.435) \color{green}$\uparrow\uparrow$ &0.709$^*$ (0.705, 0.712) \color{green}$\uparrow\uparrow$&0.236$^*$ (0.232, 0.244) \color{green}$\uparrow\uparrow$\\
         
         MSN$+ x_{icu}$  & 0.744 (0.723, 0.765) $-$ & 0.398 (0.386, 0.422) $-$ & 0.704$^*$ (0.700, 0.708) \color{green}$\uparrow\uparrow$& 0.227$^*$ (0.225, 0.235) \color{green}$\uparrow\uparrow$\\
         
         MSN$+ x_{D}$ & 0.746 (0.726, 0.767) \color{green}$\uparrow$ & 0.379 (0.368, 0.399) \color{red}$\downarrow\downarrow$  & 0.713$^*$ (0.709, 0.717) \color{green}$\uparrow\uparrow$& 0.238$^*$ (0.235, 0.245) \color{green}$\uparrow\uparrow$\\ 
         
         MSN$+ x_{SM}$   & 0.766 (0.745, 0.786) \color{green}$\uparrow\uparrow$ & 0.402 (0.387, 0.427) \color{green}$\uparrow$ & \color{blue}{0.716$^*$ (0.712, 0.719)} \color{green}$\uparrow\uparrow$& \color{blue}{0.241$^*$ (0.236, 0.249)} \color{green}$\uparrow\uparrow$ \\
         
         MSN$+ x_{SI}$   & 0.765 (0.745, 0.783) \color{green}$\uparrow\uparrow$ & 0.387 (0.375, 0.407) \color{red}$\downarrow$ & 0.712$^*$ (0.708, 0.715) \color{green}$\uparrow\uparrow$& 0.234$^*$ (0.231, 0.241) \color{green}$\uparrow\uparrow$\\
         
         MSN$+ x_{D+SM}$   & 0.752 (0.727, 0.778) \color{green}$\uparrow$ & 0.391 (0.379, 0.423) \color{red}$\downarrow$ & 0.704$^*$ (0.700, 0.707) \color{green}$\uparrow\uparrow$& 0.227$^*$ (0.224, 0.234) \color{green}$\uparrow\uparrow$ \\
         
         MSN$+ x_{D+SI}$   & 0.755 (0.738, 0.777) \color{green}$\uparrow\uparrow$ & 0.388 (0.376, 0.408) \color{red}$\downarrow$ & 0.709$^*$ (0.705, 0.712) \color{green}$\uparrow\uparrow$& 0.233$^*$ (0.230, 0.241) \color{green}$\uparrow\uparrow$\\
         
         MSN$+ x_{SM+D}$   & 0.769 (0.752, 0.794) \color{green}$\uparrow\uparrow$ & 0.406 (0.403, 0.432) \color{green}$\uparrow$  & 0.708$^*$ (0.704, 0.711) \color{green}$\uparrow\uparrow$& 0.236$^*$ (0.232, 0.243) \color{green}$\uparrow\uparrow$\\
         
         MSN$+ x_{D+SM+SI}$   & \color{blue}{0.770$^{\dagger}$ (0.746, 0.781)} \color{green}$\uparrow\uparrow$ & 0.409 (0.403, 0.454) \color{green}$\uparrow$  & 0.711$^*$ (0.706, 0.716) \color{green}$\uparrow\uparrow$& 0.233$^*$ (0.230, 0.240) \color{green}$\uparrow\uparrow$ \\
         \hline
        \multicolumn{5}{l}{{\color{black}\small $^*$  Statistical significance results with respect to vanilla MSN ($p < 0.001$)}} \\
        \multicolumn{5}{l}{{\color{black}\small $^{\dagger}$  Statistical significance results with respect to vanilla MSN ($p < 0.01$)}} \\
    \end{tabular}}
    \label{tab:ext-val-t-1}
\end{table}

\begin{table}[!ht]
    \centering
    \caption{\small{\textbf{Performance results of external validation for fine tuning of self-supervised methods using ViT-T as backbone model}. We summarize AUROC and AUPRC results on CheXpert and NIH-14 test sets including 95\% confidence intervals. The best results are shown in blue. We use {\color{green}$\uparrow$} and {\color{red}$\downarrow$} to indicate whether the performance of a given model is 0.5-1.5\% better or worse than the reference model (i.e., vanilla MSN).    {\color{green}$\uparrow\uparrow$} and {\color{red}$\downarrow\downarrow$} indicate whether the difference is 1.5\% better or worse than the reference model, respectively. $-$ indicates that the performance difference is less than 0.5\% compared to the reference model.}}
    \resizebox{0.8\linewidth}{!}
    {
    \begin{tabular}{l l l  l l }
        \hline
         &\multicolumn{2}{c}{\textbf{CheXpert}} & \multicolumn{2}{c}{\textbf{NIH-14}} \\\hline
         &\textbf{AUROC (CI)} & \textbf{AUPRC (CI)} & \textbf{AUROC (CI)} & \textbf{AUPRC (CI)} \\\hline
         
         MSN & \color{blue}{0.854 (0.834, 0.878)} & 0.494 (0.491, 0.541)  & 0.773 (0.769, 0.777) & 0.330 (0.323, 0.342) \\\hline
         
         MSN$+ x_{sex}$  & 0.842 (0.819, 0.854) \color{red}$\downarrow$ & \color{blue}{0.510 (0.499, 0.540)} \color{green}$\uparrow$ & 0.768 (0.763, 0.772) \color{red}$\downarrow$ &0.325 (0.318, 0.335) \color{red}$\downarrow$ \\
         
         MSN$+ x_{age}$  & 0.846 (0.820, 0.864) \color{red}$\downarrow$&0.492 (0.467, 0.519) $-$ & 0.772 (0.768, 0.776) $-$ & 0.330 (0.324, 0.341) $-$ \\
         
         MSN$+ x_{view}$ & 0.835 (0.821, 0.849) \color{red}$\downarrow\downarrow$ & 0.488 (0.452, 0.514) \color{red}$\downarrow$ & 0.776 (0.771, 0.780) $-$ &0.335 (0.328, 0.344) \color{green}$\uparrow$\\
         
         MSN$+ x_{pos}$  & 0.839 (0.819, 0.860) \color{red}$\downarrow$ &0.495 (0.480, 0.533) $-$ & 0.768 (0.764, 0.773) \color{red}$\downarrow$ &0.329 (0.322, 0.341) $-$ \\
         
         MSN$+ x_{mort}$ & 0.822 (0.806, 0.845) \color{red}$\downarrow\downarrow$ & 0.486 (0.484, 0.514) \color{red}$\downarrow$ & 0.772 (0.767, 0.776) $-$ & 0.326 (0.319, 0.337) $-$ \\
         
         MSN$+ x_{icu}$  & 0.832 (0.810, 0.853) \color{red}$\downarrow\downarrow$ & 0.465 (0.453, 0.538) \color{red}$\downarrow\downarrow$ & 0.769 (0.765, 0.773) $-$ &0.332 (0.325, 0.343) $-$  \\
         
         MSN$+ x_{D}$ & 0.822 (0.811, 0.842) \color{red}$\downarrow\downarrow$ & 0.478 (0.466, 0.526) \color{red}$\downarrow\downarrow$ & 0.768 (0.763, 0.773) \color{red}$\downarrow$ &0.330 (0.323, 0.341) $-$ \\ 
         
         MSN$+ x_{SM}$   &0.824 (0.805, 0.831) \color{red}$\downarrow\downarrow$ & 0.474 (0.468, 0.501) \color{red}$\downarrow\downarrow$ & 0.773 (0.768, 0.777) $-$ &0.335 (0.328, 0.345) $-$ \\
         
         MSN$+ x_{SI}$   &0.829 (0.800, 0.858) \color{red}$\downarrow\downarrow$ & 0.491 (0.453, 0.527)$-$ & 0.772 (0.768, 0.775) $-$ &0.330 (0.324, 0.340) $-$ \\
         
         MSN$+ x_{D+SM}$   &0.832 (0.804, 0.858) \color{red}$\downarrow\downarrow$ &0.494 (0.475, 0.535) $-$ & 0.767 (0.763, 0.771) \color{red}$\downarrow$ &0.318 (0.313, 0.327) \color{red}$\downarrow$ \\
         
         MSN$+ x_{D+SI}$   & 0.838 (0.815, 0.848) \color{red}$\downarrow\downarrow$ & 0.482 (0.460, 0.515) \color{red}$\downarrow\downarrow$ & \color{blue}{0.775 (0.770, 0.779)} \color{green}$\uparrow$ & \color{blue}{0.337 (0.330, 0.349)} \color{green}$\uparrow$ \\
         
         MSN$+ x_{SM+D}$  &0.840 (0.819, 0.846) \color{red}$\downarrow$ &0.496 (0.484, 0.536) $-$ & 0.772 (0.768, 0.777) $-$ & 0.334 (0.328, 0.346) $-$ \\
         
         MSN$+ x_{D+SM+SI}$ & 0.846 (0.836, 0.866) \color{red}$\downarrow$ & 0.492 (0.469, 0.531) $-$ &0.772 (0.767, 0.776) $-$ &0.334 (0.328, 0.344) $-$ \\
         \hline
         
    \end{tabular}}
    \label{tab:ext-val-t-2}
\end{table}

\end{document}